\documentclass[12pt]{iopart}
\usepackage{graphicx}
\graphicspath{ {Images/}}
\usepackage{subfig}

\begin{document}

\title[A deep learning classification scheme to segment OARs in brain cancer patients]{A deep learning classification scheme based on augmented-enhanced features to segment organs at risk on the optic region in brain cancer patients}

\author{J Dolz$^{1,2}$, N Reyns$^{2,3}$, 
N Betrouni$^2$, D Kharroubi$^2$, M Quidet$^2$, L Massoptier$^1$, and M Vermandel$^{2,3}$}

\address{$^1$ AQUILAB, Loos-les-Lille, France}
\address{$^2$ University of Lille, Inserm, CHU Lille, U1189 ONCO-THAI-Image Assisted Laser Therapy for Oncology, F-59000 Lille, France}
\address{$^3$ Neurosurgery Department, University Hospital Lille, Lille, France}
\ead{jose.dolz.upv@gmail.com}
\vspace{10pt}

\begin{abstract}
Radiation therapy has emerged as one of the preferred techniques to treat brain cancer patients. During treatment, a very high dose of radiation is delivered to a very narrow area. Prescribed radiation therapy for brain cancer requires precisely defining the target treatment area, as well as delineating vital brain structures which must be spared from radiotoxicity. Nevertheless, delineation task is usually still manually performed, which is inefficient and operator-dependent. Several attempts of automatizing this process have reported, however, marginal results when analyzing organs in the optic region. In this work we present a deep learning classification scheme based on augmented-enhanced features to automatically segment organs at risk in the optic region -optic nerves, optic chiasm, pituitary gland and pituitary stalk-. Fifteen MR images with various types of brain tumors were retrospectively collected to undergo manual and automatic segmentation. Mean Dice Similarity coefficients of 0.79, 0.83, 0.76 and 0.77, respectively, were reported in this study. Incorporation of proposed features yielded to improvements on the segmentation with respect to classical features. Compared with support vector machines, our method achieved better performance with less variation on the results, as well as a considerably reduction on the classification time. Performance of the proposed approach was also evaluated with respect to manual contours. In this case, results obtained from the automatic contours mostly lie on the variability of the observers. Additionally, in cases where our method was under performing with respect to manual raters, statistical analysis showed that there were not significant differences between them. These results suggest therefore that the proposed system is more accurate than other presented approaches, up to date, to segment these structures. The speed, reproducibility, and robustness of the process make the proposed deep learning-based classification system a valuable tool for assisting in the delineation task of small OARs in brain cancer.
\end{abstract}

%
\vspace{2pc}
\submitto{Journal of Physics in Medicine and Biology}
%
%
%

\section{Introduction}
\label{intro}

During radiation treatment planning (RTP) high doses are delivered into very narrow areas. Although techniques have evolved along the years and have been largely improved, radiation is still spread beyond the target. To constrain the risk of severe toxicity of critical brain structures, i.e. the organs at risk (OARs), volume measurements and localization of these structures are required. Among available image modalities, magnetic resonance (MR) images are extensively used to segment most of the OARs. 

Nevertheless, manual delineation of brain structures could be prohibitively time-consuming, prone to error, operator dependent and a poorly reproducible process \cite{Bondiau,Deeley}. Thus, image segmentation has become a central part in the RTP, being often a limiting step of it. Automatic segmentation algorithms are therefore highly recommended in order to surmount such disadvantages. There are, however, several technical difficulties that make of the task of automatic segmentation of OARs from MR images in brain cancer patients still a challenging problem.


There exist a number of atlas-based approaches that have already attempted to segment some OARs and brain structures in patients undergoing radiotherapy \cite{Bondiau,Deeley,dhaese,isambert,noble2011atlas,conson2014automated}. Most of the available techniques need to combine CT and MR sequences to accomplish the task. Because CT and MR images are not always acquired simultaneously, combining both sequences might incorporate an additional step into the chain, in which images from the same patient are aligned, prior to apply the segmentation algorithm. Although atlas-based approaches have been reported to produce good results for most head structures \cite{dolz2015Review}, limited success has been achieved when segmenting organs near the optic region, such as the optic nerves, chiasm or the pituitary gland and stalk \cite{Bondiau,isambert}. These structures are particularly difficult to segment mainly due to lack of contrast in some regions, heterogeneities in the texture in some of them, complexity of shape and/or shape and location variability across patients.

As alternative to atlas-based methods, Bekes et al. \cite{bekes2008geometrical} proposed a geometrical model-based segmentation technique. In addition to optic chiasm and nerves, the eyes and lenses were also included in the evaluation. Whilst segmentation of eyes and lenses were satisfactory, segmentation of optic nerves and chiasm was below their expectations. Repeatability and reproducibility of the automatic results made the method not being usable for RTP for these two challenging structures. On the other hand, success on segmenting the pituitary gland and stalk has been even more limited, with very few works having reported any result \cite{Deeley,isambert}.

Inspired by the recent success of deep learning in the fields of computer vision and medical imaging, we considered its use in the presented work as alternative to segment OARs in the optic region. Deep learning has revived during last years, and deep networks have already been used on MR brain images, with special focus on segmentation of tumor \cite{pereira2016brain} and some brain structures \cite{dolz20163d}. With regards to segmentation of OARs in brain cancer, few attempts have been so far presented \cite{kim2013unsupervised,guo2014segmenting}, where hippocampal segmentation was addressed. Among all the deep learning approaches, convolutional neural networks (CNNs) have demonstrated to be very powerful in In biomedical imaging. In these networks, two or three-dimensional patches are commonly fed into the deep network. A hierarchical representation of the input data is then learned, decoding the important information contained on the data. By doing this, a deep network is able to ensure discriminative power for the learned features. However, valuable information inherited from classical machine learning approaches to segment brain structures is not included into these convolutional architectures. This knowledge may come in the form of likelihood voxel values, voxel location, as well as textural information, for example, which are greatly useful to segment structures that share similar intensity properties. Networks based on convolutional filters, i.e. CNN, perfectly suit to deal with data presenting a grid structured representation, such as 2D or 3D image patches. However, when input data composed by features not presenting a grid-based representation is employed, CNNs might not represent the best solution. Because we wish to employ arrays composed by concatenation of different features we consider the use of denoised auto encoders (DAE) instead, which has demonstrated to be able to deal with such type of features arrays \cite{DolzCMIG}. Another reason for employing DAE is because of the limited size of the number of training and labeled data. Despite the different strategies to network weights initialization, if not enough training is available there exist a high risk of overfitting. DAEs act as a pre-training step, obtaining an approximate initialization of the weights in an unsupervised fashion. Thanks to this the network can be trained with such limited amount of data while avoiding overfitting.

In this paper, we present a classification system based on a deep learning technique to segment OARs in the optic region. Instead of using image patches, we use a pile of hand-crafted features as input of the network. In addition to features typically employed in machine learning approaches to segment brain structures \cite{Dolz2014,Dolz2015CARS,Powell}, we propose the extension of the features vector to improve voxel characterization. The novel augmented enhanced features vector (AE-FV) incorporates more information about a voxel and its environment, such as contextual features, and first order statistics and spectral measures. This allows to successfully segment more complex structures, such as the optic nerves, for example. In addition, clinical evaluation of our automatic system involving manual segmentations from several experts is also assessed in our experiment.

\section{Methods and materials}
\label{sec:methods}

\subsection{Composition of the augmented enhanced features vector (AE-FV)}
\label{ssec:featuresextraction}


For most existing machine learning segmentation methods, the features vector for a voxel \textit{v} is composed by: voxel intensities in an image patch centered at \textit{v}, likelihood of \textit{v} of belonging to a particular structure, and location of \textit{v} \cite{Dolz2015CARS,Powell}. In addition to classical features, we incorporate gradient patch information, contextual features and first order statistics and spectral measures for each voxel \textit{v}.

\subsubsection{Classical machine learning features}
\label{sssec:remainFeat}

There are a number of features that have been successfully employed to segment some brain structures in machine learning based approaches, which are common in many approaches. Image intensities are among these commonly employed features. Intensities can be used either as a patch around the voxel under examination \cite{Powell} or as a set of voxels in a specific and meaningful direction \cite{Dolz2015CARS}. To complete the group of classical features, likelihood of a voxel \textit{v} of belonging to a particular structure, and location of \textit{v} have been also largely employed \cite{DolzCMIG,Dolz2014,Dolz2015CARS,Powell}. 


\subsubsection{Gradient and contextual features}
\label{sssec:gradContextFeat}
The term of augmented features, and the inclusion of gradient and contextual features into the features vector, was already introduced by \cite{bai2015multi}. In their work, gradient orientations of all the voxels on each patch were used. Following their work, to describe relative relations between an image patch and its surroundings, contextual features are used. For each voxel \textit{v}, a number of regions around its surroundings are sampled, radiating from voxel \textit{v} with equal degree intervals and at different radius. To obtain a continuous description of the context, intensity difference between the voxel \textit{v} and a patch \textit{P} is defined:

\begin{equation}
	d_{\textit{v},P} = \mu_{P}-I_{\textit{v}}
\end{equation}

where $\mu_{P}$ is the mean intensity of the patch \textit{P} and $I_{\textit{v}}$ is the intensity of the voxel \textit{v}. In addition, a compact and binary context description is obtained by employing a descriptor known as BRIEF \cite{calonder2012brief}:


\begin{equation}
	b_{\textit{v},P }=\{ 1\enspace if \enspace  I_{\textit{v}} < \mu_{P},\enspace  0 \enspace   otherwise\}
\end{equation} 

Then, for each patch, the contextual feature includes both the continuous and binary descriptor for all the neighbor regions sampled.  

\subsubsection{Features from texture analysis}
\label{sssec:featText}

Texture analysis (TA) has proven to be a potentially valuable and versatile tool in neuro MR imaging \cite{kassner2010texture}. TA can be divided into several categories according to the means employed to evaluate the inter-relationships of the pixels. Statistical methods are the most widely used in medical images. On these methods, the spatial distribution of grey values is analyzed by computing local features at each point in the image, and deriving a set of statistics from the distributions of the local features. Local features are defined by the combination of intensities at specific position relative to each point in image. In the literature, the use of these features to characterize textures have been mainly employed for classification of images \cite{aggarwal2012first} or for the characterization of healthy and pathological human cerebral tissues\cite{qurat2010classification}. Nevertheless, their use as discriminant factor in the segmentation of critical structures in brain cancer has not been investigated yet.

To quantitatively describe the first order statistical features of an image patch \textit{P}, the following image features obtained from the histogram were employed: mean, variance, skewness, kurtosis and entropy. The mean takes the average level of intensity of the patch \textit{P}, whereas the variance describes the variation of intensity around the mean. Skewness is a measure of symmetry, or more precisely, the lack of symmetry. The skewness for a normal distribution is zero, and any symmetric data should have a skewness near zero. Kurtosis is a measure of whether the data are peaked or flat relative to a normal distribution. That is, data sets with high kurtosis tend to have a distinct peak near the mean, decline rather rapidly, and have heavy tails. Data sets with low kurtosis tend to have a flat top near the mean rather than a sharp peak. A uniform distribution would be the extreme case. 

%

\begin{table}[]
\centering
\scriptsize
\begin{tabular}{|l|l|c|}
\hline
\textbf{Features set name} & \textbf{Features included}                                                                                                                                                                                                           & \multicolumn{1}{l|}{\textbf{Vector size}} \\ \hline
\textbf{Classical}         & \begin{tabular}[c]{@{}l@{}}Intensity of voxel under examination\\ Intensity of voxel neighborhood (3D)\\ Probability voxel value\\ Spherical voxel coordinates\\ Intensity of 8 voxels along maximum gradient direction\end{tabular} & 137                                       \\ \hline
\textbf{Augmented}         & \begin{tabular}[c]{@{}l@{}}Classical\\ Gradient Patch in 2D \\ (Horizontal and vertical magnitudes and orientation)\\ Contextual Features\end{tabular}                                                                               & 276                                       \\ \hline
\textbf{Textural}          & \begin{tabular}[c]{@{}l@{}}Classical\\ Mean \\ Variance\\ Entropy\\ Energy \\ Kurtosis\\ Skewness\\ Wavelet patch decomposition\end{tabular}                                                                                                   & 147                                       \\ \hline
\textbf{AE-FV}             & \begin{tabular}[c]{@{}l@{}}Classical\\ Augmented (except classical)\\ Textural (except classical) \end{tabular}                                                                                                                                                             & 286                                    \\ \hline
\end{tabular}
\caption{Features sets.}
\label{tab:features}
\end{table}

Statistical based features may lack the sensitivity to identify larger scale or more coarse changes in spatial frequency. To evaluate spatial frequencies at multiple scales wavelet functions can be employed \cite{mallat1989theory}. The basic idea of the algorithm is to divide the input images into respective a hierarchy of sub-bands with sequential decrease in resolution. In the medical field, a major usage of Discrete wavelet transform (DWT) has been mainly noticed for classifying MR brain images into normal and abnormal tissue \cite{john2012brain}, not being fully exploited yet in image segmentation. 

\subsection{Deep Learning based classification scheme}
\label{ssec:scheme}

Voxel classification is assessed by using a deep learning technique, which is known as Stacked Denoised auto-encoders (SDAE). This technique learns hierarchical correlations between feature representations in a given dataset through a semi-supervised learning approach \cite{Bengio2009}. Unlike common deep learning based approaches, which employ image patches as input of the network, an array composed of hand crafted features is used as input instead. Hence, the proposed approach follows a hybrid architecture which unsupervisedly learns a compact representation of the hand-crafted features followed by a supervised fine tuning of the parameters of the network.

\subsubsection{Denoising Auto-Encoder (DAE)}
\label{sssec:denoisAE}

In its simplest representation, an auto-encoder (AE) is composed by two components: an encoder \textit{h($\cdot$)} and a decoder \textit{g($\cdot$)}. While the encoder maps the input \textit{x} $\in$ \textit{R$_d$} to some hidden representation \textit{h(x)} $\in$ \textit{R$_dh$}, the decoder maps the hidden representation back to a reconstructed version of the input \textit{x}, so that \textit{g(h(x))}~$\approx$~\textit{x}. An AE is therefore trained to minimize the discrepancy between the data and its reconstruction. Nevertheless, if no other constraint besides reconstruction error minimization is imposed, it might potentially happen that an AE just learns the identity function. This effect would lead to simply copy the input, for which many encodings would be useless, making the AE not capable to differentiate test examples from other input configurations. One solution to prevent this is to add randomness in the transformation from input to reconstruction, which is exploited in Denoising Auto-Encoders (DAEs) \cite{Vincent2008,Vincent}. 

The Denoising Auto-Encoder (DAE) is typically implemented as a one-hidden-layer neural network which is trained to reconstruct a data point \textit{x} $\in$~$\Re$~$^D$ from its corrupted version $\tilde{x}$ \cite{Vincent2008}. This leads to a partially destroyed version $\tilde{x}$ by means of a stochastic mapping $\tilde{x}\sim q_D(\tilde{x}|x)$. Therefore, to convert an AE class into a DAE class, only adding a stochastic corruption step that modifies the input is required, which can be done in many ways. In this work, for example, the stochastic corruption process consists in randomly setting some of the inputs to zero. A single DAE is limited in regards what it can represent, because it is simply a shallow model in terms of learning. Therefore, several DAEs are stacked to form a deep network by feeding the hidden representation of the DAE found on the layer below as input to the current layer~\cite{Vincent}.

\subsubsection{Network training.}
\label{sssec:netTraining}

Weights between layers of the network are initially learned via the unsupervised pre-training step, which is done greedily, i.e. one layer at a time. Each layer is trained as a DAE by minimizing the reconstruction of its input. The high level DAE uses the output of the lower level DAE as input. Once the first \textit{k} layers are trained, the (\textit{k}+1)$^{th}$ layer can be trained because the latent representation from the layer below can be then computed. Once all the weights of the network are unsupervisedly computed, a logistic regression layeris added on top of the encoders, yielding a deep neural network amenable to supervised learning. Thus, the network goes through a second stage of training called \textit{fine-tuning}, where prediction error is minimized on a supervised task \cite{Vincent}. A gradient-based procedure such as stochastic gradient descent is employed in this stage. The hope is that the unsupervised initialization in a greedy layer-wise fashion has put the parameters of all the layers in a region of parameter space from which a good local optimum can be reached by local descent. 


\subsection{Study design and experiment set-up}
\label{ssec:experiments}
 
\subsubsection{Dataset}
\label{sssec:dataset}

MRI data from 15 patients who underwent Leksell Gamma Knife Radiosurgery were used in the present study. For each patient, optic nerves, optic chiasm, pituitary gland and pituitary stalk were manually delineated by three experts trained and qualified for radiosurgery delineation. Protocol for delineation was described before contouring session and it followed the RTOG guidelines. To achieve Dicom RT contouring structures, Artiview \textregistered 3.0 (Aquilab) was used after a training session. Average time of manual contouring was 7 min and 34 s ($\pm$2 min and 53 s), 1 min and 52 s ($\pm$38 s), 3 min and 8 s ($\pm$55 s) and 2 min and 41 s ($\pm$49 s) for optic nerves, optic chiasm, pituitary gland and pituitary stalk, respectively. Two different MRI facilities were used to acquire images according to the radiosurgery planning protocol (Table \ref{table:devices}). 

\begin{table}[h]
\footnotesize
\begin{tabular}{l|cccccc}
\hline
MRI System              & TE(ms) & TR(ms) & \begin{tabular}[c]{@{}c@{}}Echo \\ number\end{tabular} & Matrix size & Seq. Name & \begin{tabular}[c]{@{}c@{}}Voxel Size \\ (mm$^3$)\end{tabular} \\ \hline
\begin{tabular}[c]{@{}l@{}}Philips Achieva \\1.5T \end{tabular}   & 4.602  & 25     & 1           & 256x256     & T1 3D FFE & 1x1x1            \\
\begin{tabular}[c]{@{}l@{}}GEHC Optima\\ MR450w 1.5T\end{tabular} & 2.412  & 5.9    & 1           & 256x256     & FSPGR     & 0.8203x0.8203x1          \\ \hline
\end{tabular}
\caption{Acquisition parameters on the 2 MRI devices.}
\label{table:devices}
\end{table}

To conduct a validation analysis of the quality of image segmentation, it is typically necessary to know a voxel-wise reference standard. Nevertheless, image segmentation in the medical domain often lacks from a universal known ground truth. Even though a single manual rater provides realistic data, contours may suffer from intra- and inter-observer variability. Thus, a number of observers and target patients that provide a good statistical analysis is often required. Accordingly, this study has been designed to quantify variation among clinicians in delineating OARs and to assess our proposed classification scheme in this context. Therefore, available manual contours from the experts were used to create the simulated ground truth. Reference contours have been obtained by using the computationally simple concept of the voting rule approach. In this technique, each voxel of the simulated ground truth is mapped to a given class the most frequent class in corresponding voxels of manual annotations. Due to differences between observers and the constrained size of our dataset, generated ground truth could not always be satisfactory and might be considered as corrupted data, particularly if they are employed for learning. To ensure this does not happen, an external expert reviewed the generated ground truth and performed small modifications, if needed.

\subsubsection{Training and classification schemes.}
\label{sssec:ParametersSetting}

\begin{figure}[h!]
\centering
\begin{tabular}{cc}
\subfloat[Training scheme]{\includegraphics[width=0.5\linewidth]{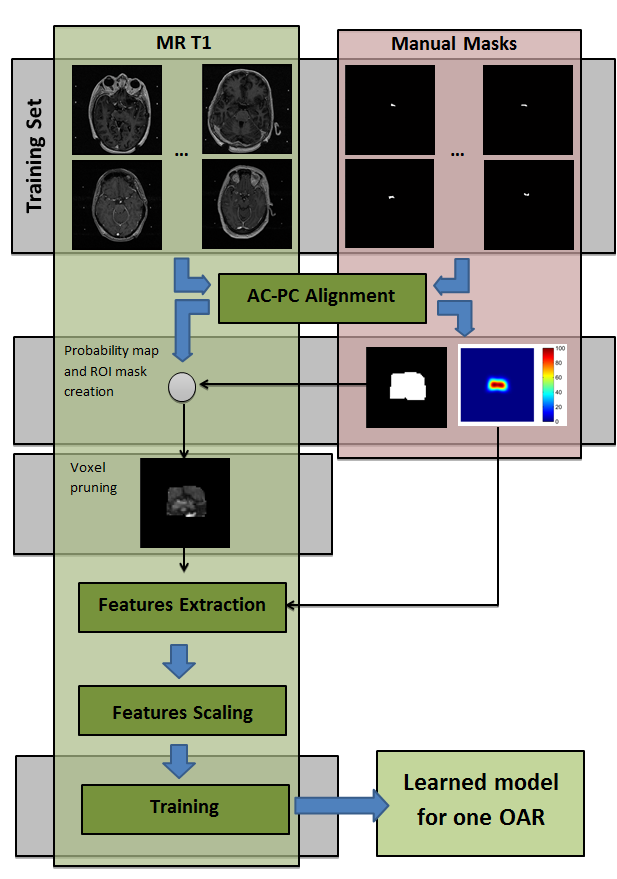}} & 
\subfloat[Classification scheme]{\includegraphics[width=0.5\linewidth]{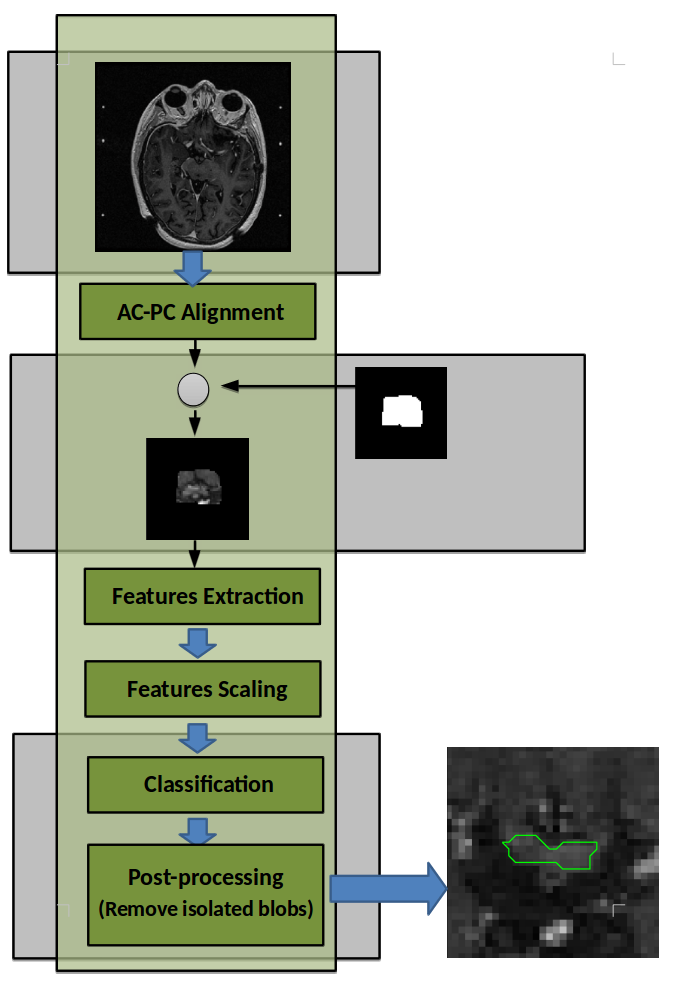}} \\ 
\end{tabular}
\caption{Training (a) and classification (b) workflow of the proposed system. }
\label{fig:trainClass}
\end{figure}

Figure \ref{fig:trainClass} shows the training (a) and classification (b) workflow. In the proposed approach, and as in~\cite{Powell,Dolz2015CARS}, MR T1 images and manual OARs delineations were spatially aligned such that the anterior commissure and posterior commissure (AC$-$PC) line was horizontally oriented in the sagittal plane, and the inter hemispheric fissure was aligned on the two other axes. This step was therefore necessary to initialize the segmentation for a new target patient. In addition, images which resolution differed from 1 x 1 x 1 mm$^3$ were resampled to this resolution. Once images were aligned, the process of extracting the features, which is detailed in next section, was carried out. Next step was either training the network or performing the classification. In the former case, the output of the system was the learned model for one OAR. This means that for each of the OARs the whole process, with exception of the AC-PC alignment, was repeated. In the case of classification, an additional post-processing step, where small isolated blobs are removed is included.

%
%

\subsubsection{Parameters setting.}
\label{sssec:ParametersSetting}

\subparagraph{Features extraction.}

A spatial probabilistic distribution map (SPDM) for each of the OARs was used as one of the components of the features vector. SPDM represents the probability of a voxel to belong to a given organ, and it is obtained by summing all the manual labels contained in the training data set. The resulted map is then smoothed by using a Gaussian filter with a kernel size of 3x3x3. To reduce the number of input samples that contain consistent information, the voxel space was first binarized by setting its values greater than 0.005 to 1, and the others to 0. Then, a dilation operation with a square kernel type of size 3x3x3 was applied over the binary image. Only those voxels that belonged to the inner part of the dilated image were kept to feed the prediction algorithm. Both the SPDM and the binary region of interest are shown in the step immediately after to the AC-PC alignment in figure \ref{fig:trainClass}.

MR T1 sequence was the only image modality used. For features related with intensity and gradient values, patches around each voxel of size 5x5x5 and 5x5x1 voxels were used, respectively. For the contextual features, and as in the work of \cite{bai2015multi} regions of size 3x3x1 voxels were sampled around the voxel under examination by radiation from it at every 45º, and at four different radius: 4,8,16 and 32. By combining the continuous and the binary value at each sampled patch, this led to a total of 64 contextual features for each voxel. To compute first-order textural features, patches of size 3x3x3 were extracted around each voxel. Different patches configurations were investigated. Particularly, patches of size 7, 9 and 11 were included in the features vector. However, their inclusion did not lead to significant performance improvement, but it considerably increased the computation time to extract the features. Therefore, they have not been included in our evaluation. Regarding the use of wavelet-based features, first to fourth order high-pass components from discrete wavelet decomposition were employed. The total number of features used in each features set is shown in table \ref{tab:features}. 


\subparagraph{SDAEs network.}

Choice of network parameters was based on a \textit{k}-fold cross validation strategy employing the full dataset. Samples from the whole dataset were randomly split into \textit{k} groups of same size, with \textit{k} equal to 10. While \textit{k}-1 groups were used for training, the remaining group was employed for validation. In the current work, sample refers to a single voxel, not to a patient. The convergence of the training error during the \textit{fine-tuning} phase was monitored to select the network configuration.  Several network configurations were explored and as average, the architecture that reported best performance across the four OARs was composed by four hidden layers with 400, 200, 100 and 50 units each one, respectively. The learned representation of the input had therefore a dimensionality of 50. At the end of the last layer of DAEs a softmax layer was used as output with the sigmoid function as activation function. Mini-batch learning was followed during both unsupervised pre-training of DAEs and supervised fine-tuning of the entire network. Denoising corruption level for the DAEs was set to 0.7. For training and testing purposes, \textit{leave-one-out-cross-validation} strategy was followed. In this case, however, training samples were extracted from the 14 cases employed for training at each iteration, whereas the testing samples were extracted from patient that was independent from the training set.

\subparagraph{Implementation.}While C++ was employed for image processing and features extraction steps,
MATLAB was used for training and classification purposes, by modifying the implementation provided by \cite{palm2012prediction}.


\subsubsection{Evaluation}
\label{sssec:evaluation}

Automatic segmentations were compared with the reference manual segmentations by using several metrics sensitive to different aspects of geometry. First, we used the Dice similarity coefficient (DSC) \cite{dice1945measures}, which is defined as the ratio of twice the intersection over the sum of the two segmented results, \textit{X} and \textit{Y}:

\begin{equation}
	DSC = 2 \frac{\mid V_{expert} \bigcap V_{auto} \mid}{\mid V_{expert} \mid + \mid V_{auto} \mid} 
\end{equation}
	 
where $V_{expert}$ is the expert delineation, and $V_{auto}$ is the result segmentation of the proposed approach. The DSC measure varies between [0-1], where zero indicates no overlap while 1 indicates perfect overlap. Then, to measure volume differences between automatic and reference contours, the following formula is used

\begin{equation}	
 \Delta V(\%) = \frac{\mid V_{auto} - V_{expert}\mid}{V_{expert}} \ast 100
 \label{VolDif}
\end{equation} 

where $V_{expert}$ represents the expert or reference delineation, and $V_{auto}$ is the outcome of the proposed segmentation approach. 

Although volume-based metrics have been broadly used to compare volume similarities, they are fairly insensitive to edge differences when those differences have a small impact on the overall volume. If a given segmentation is planned to be used in RTP, an analysis on shape fidelity of the segmentation outline is highly recommended. Any underinclusion on the OAR delineation might lead to a part of the healthy tissue exposed to radiation. Therefore, a surface distance measure (Hausdorff distance \cite{huttenlocher1993comparing}) was also used to evaluate the segmentation results.

In addition, sensitivity and specificity were also investigated. The numbers of true positive (TP), true negative (TN), false positive (FP), and false negative (FN) voxels were determined. The sensitivity, TP/(TP+FN), might be equal to 1 for a poor segmentation much bigger than the ground truth. On the other hand, the specificity, TN/(TN+FP), might be equal to 1 for a very poor segmentation that does not detect the object of interest at all. Consequently, a good segmentation system should have high sensitivity and specificity values. In order to assess the value of the deep learning based classification system, sensitivity and specificity values are computed before to apply any post-processing step on the resulted segmentation.

\begin{figure}[htb]
    \centering
	\includegraphics[width=0.5\linewidth]{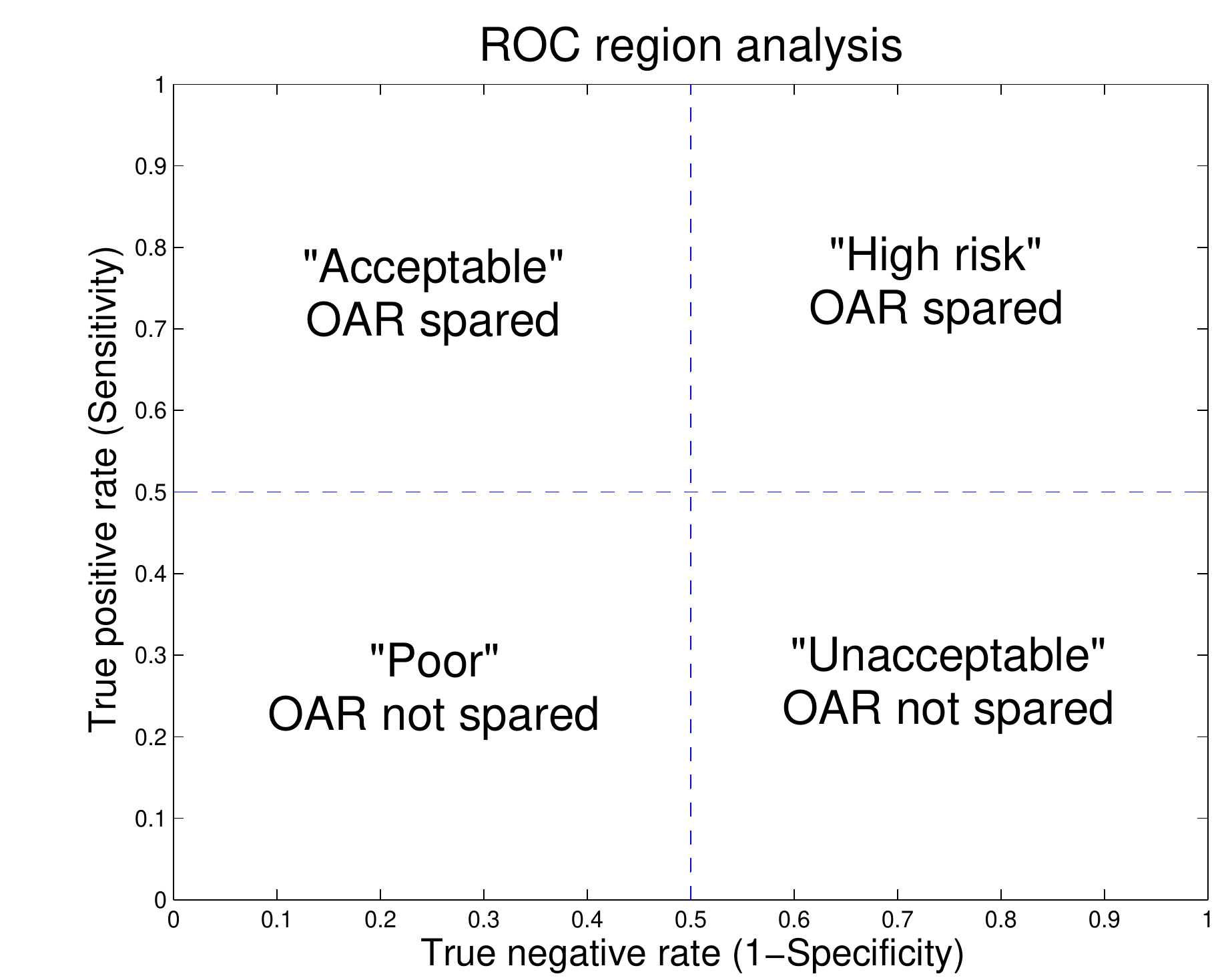}  
	\caption{ROC space sub-division to evaluate our classifier performance.}
	\label{fig:ROCArea}
\end{figure}

Receiver operating characteristic (ROC) analysis is usually employed to analyze classifiers performance. In this type of evaluation, curves defining the relation between sensitivity and (1 - specificity) are plotted. If the ROC analysis is considered from a radiotherapy point of view, FN and FP voxels must be taken into consideration when analyzing the segmentation performance. While FN voxels might lead to overirradiation of OARs voxels, FP voxels could result in a possible underirradiation of target volume voxels. Thus, the higher the sensitivity, the lower risk of overirradiation of normal tissue and the higher the specificity, the lower the risk of underirradiation of tumor tissue. Following the suggestion of \cite{andrews1985benefit}, instead of employing ROC curves to evaluate performance of a given classifier, the ROC space is used. The ROC space can be divided into four sub-spaces (Figure \ref{fig:ROCArea}). Thus, results spread over the left-top sub-space indicate acceptable contours, with the OAR spared and the PTV covered. Results lying on the right-top sub-space present a high-risk, since the OAR may be spared but with PTV not covered. Poor contours are considered when they ROC representation are present on the left-bottom sub-space. There, although the PTV is covered, it is considered that the OAR is not spared. And last, the right-bottom side of the ROC subdivision contains the unacceptable contours, with OARs not spared and PTV not covered. 
 
\section{Results}
\label{ssec:results}

\begin{figure}[t!]
\centering
\begin{tabular}{cc}
\subfloat[]{\includegraphics[width=0.475\linewidth]{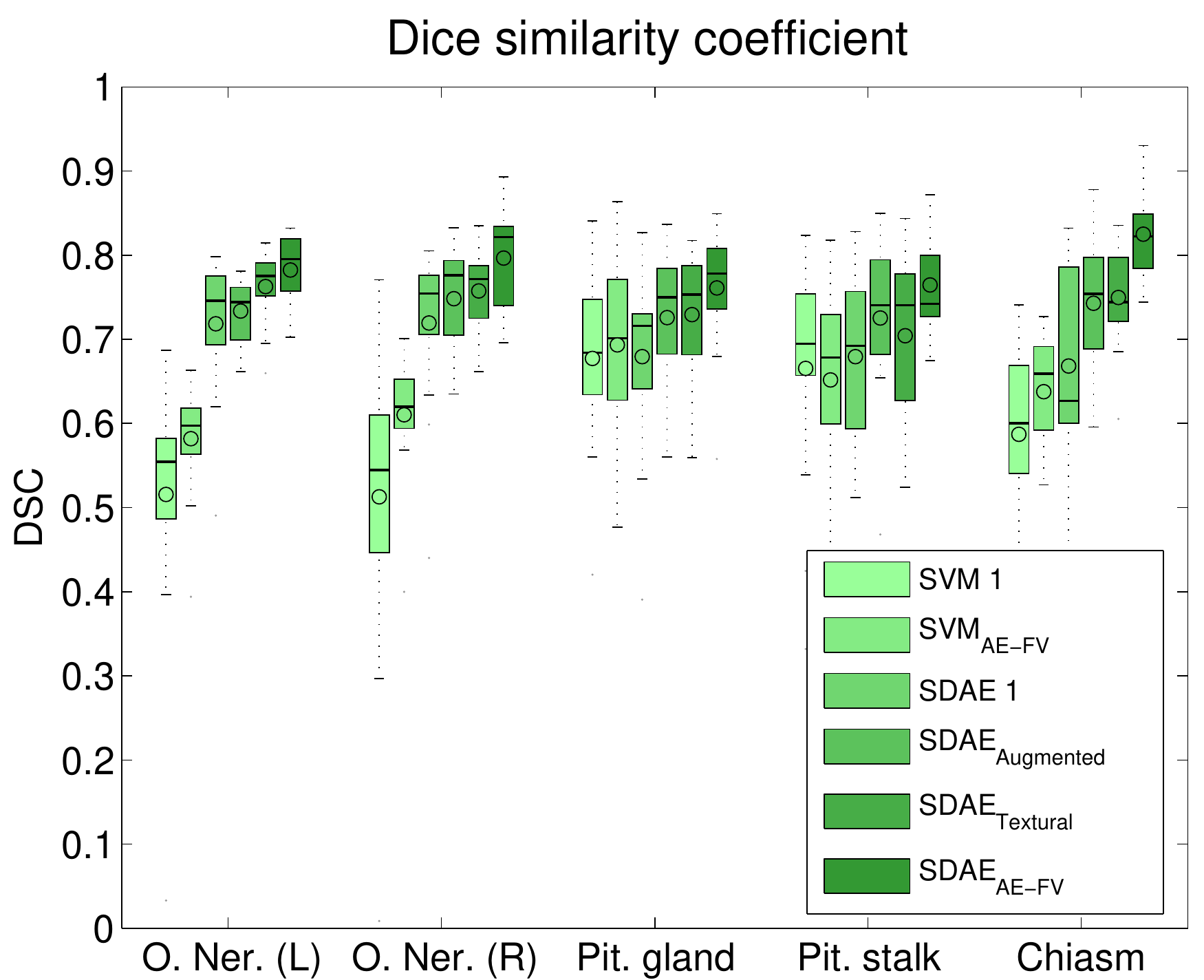}} &
\subfloat[]{\includegraphics[width=0.475\linewidth]{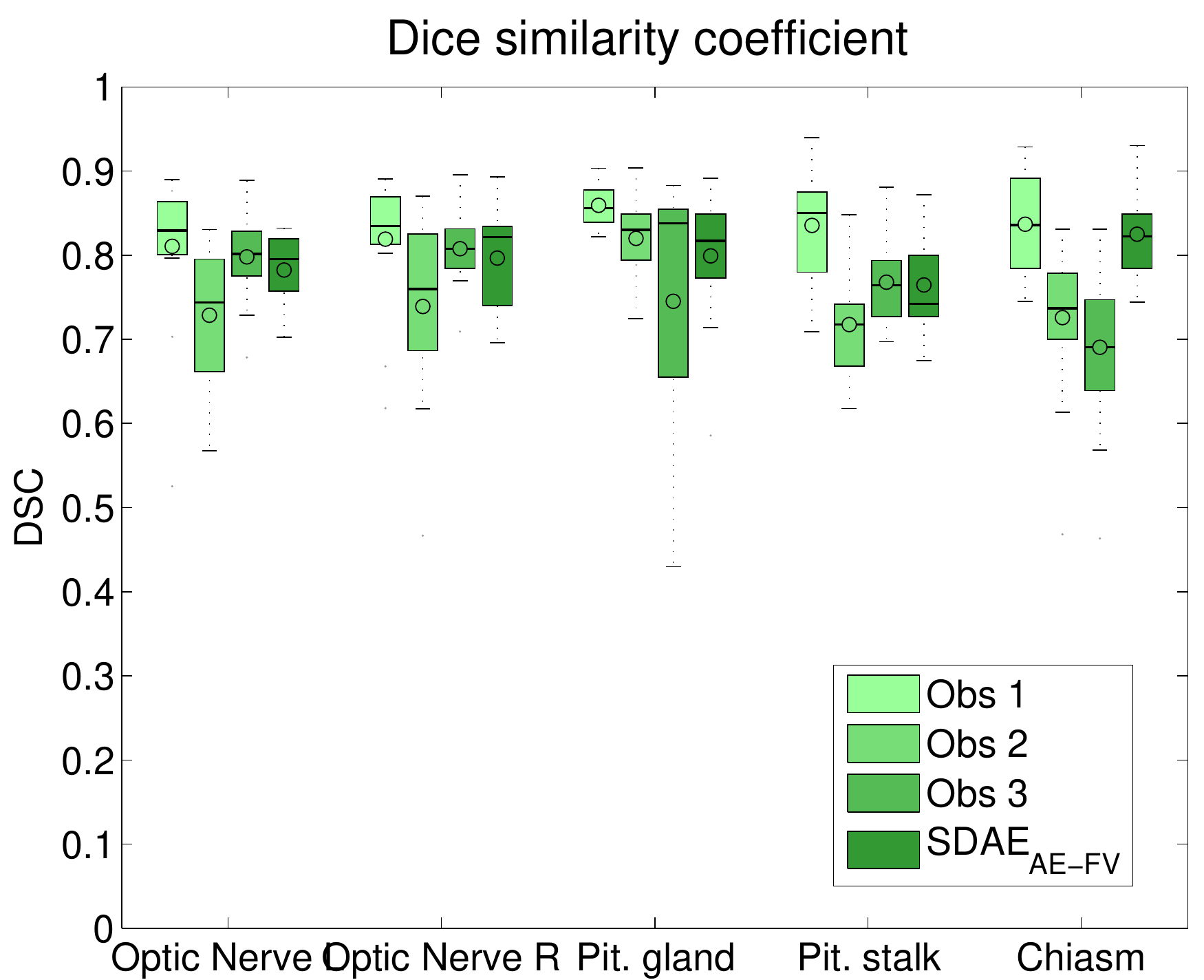}} \\
\subfloat[]{\includegraphics[width=0.475\linewidth]{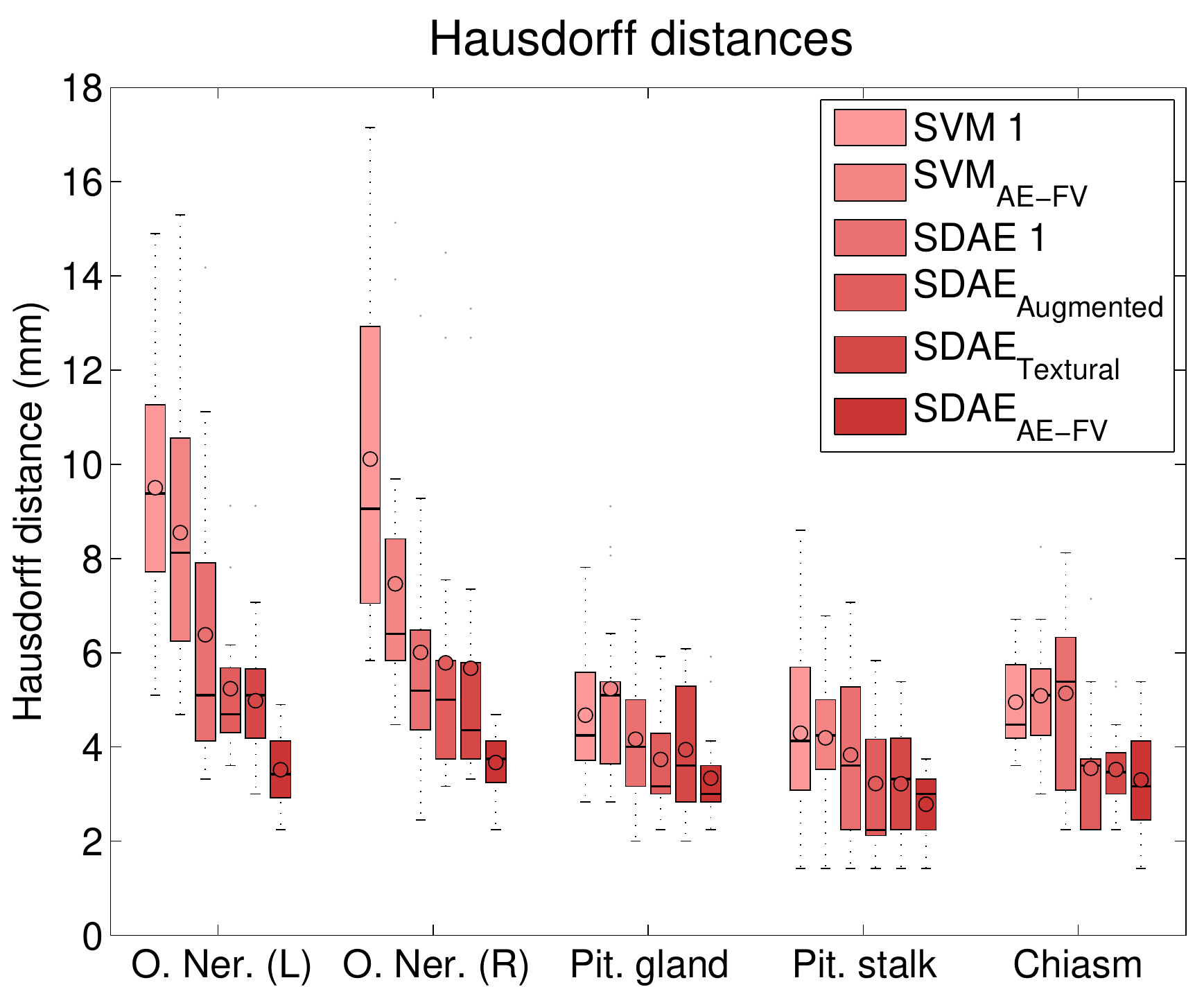}} &
\subfloat[]{\includegraphics[width=0.475\linewidth]{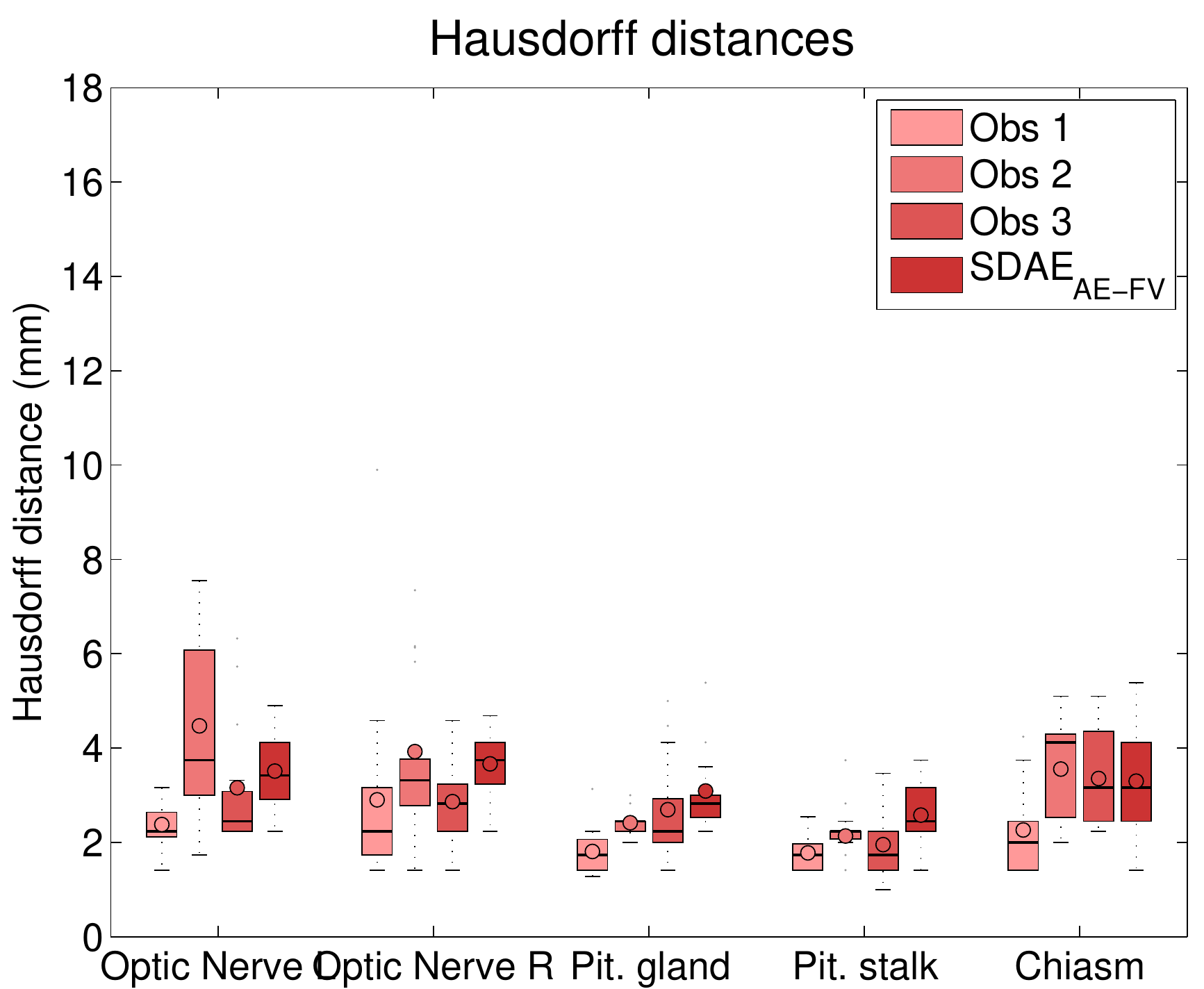}} \\
\subfloat[]{\includegraphics[width=0.475\linewidth]{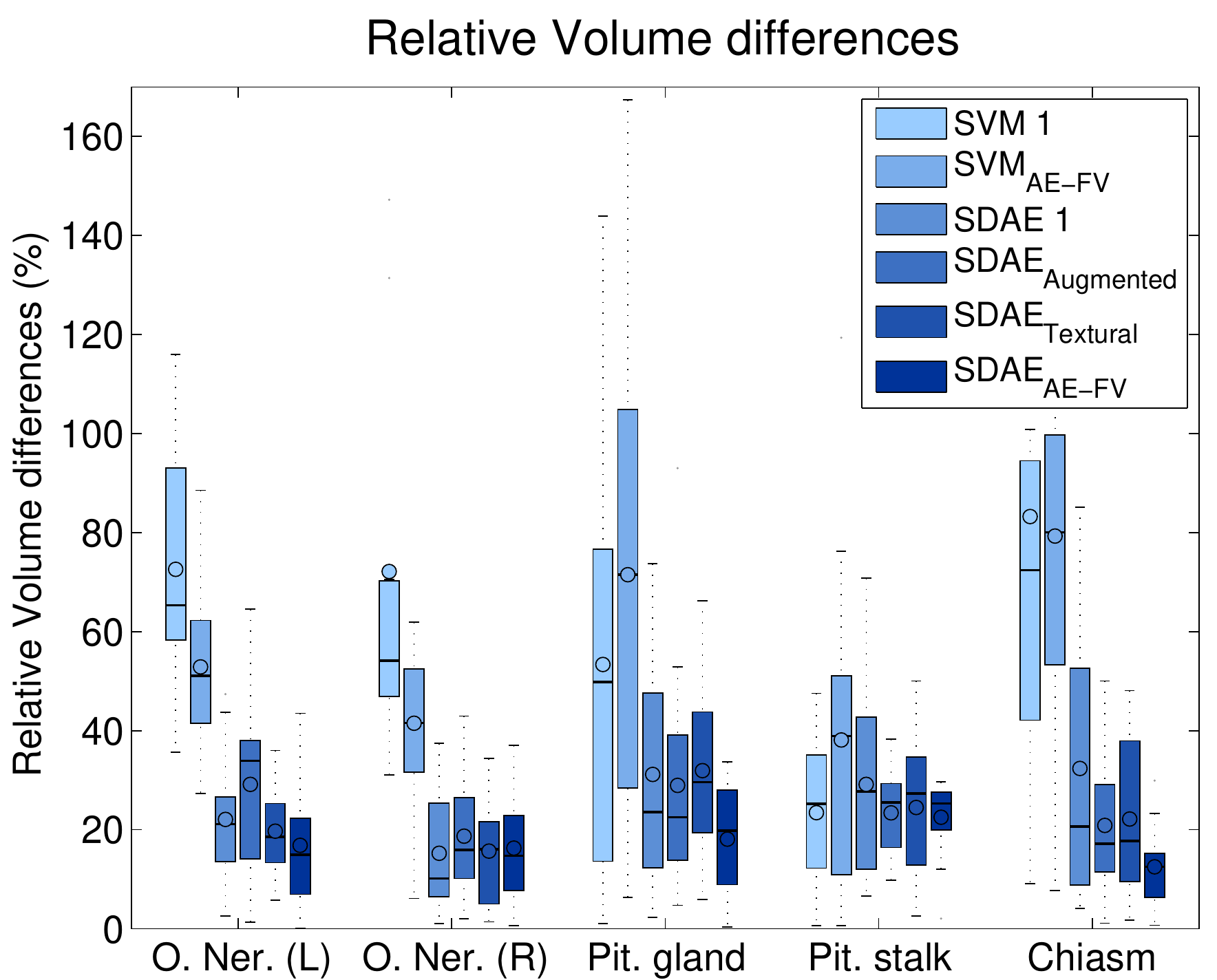}} &
\subfloat[]{\includegraphics[width=0.475\linewidth]{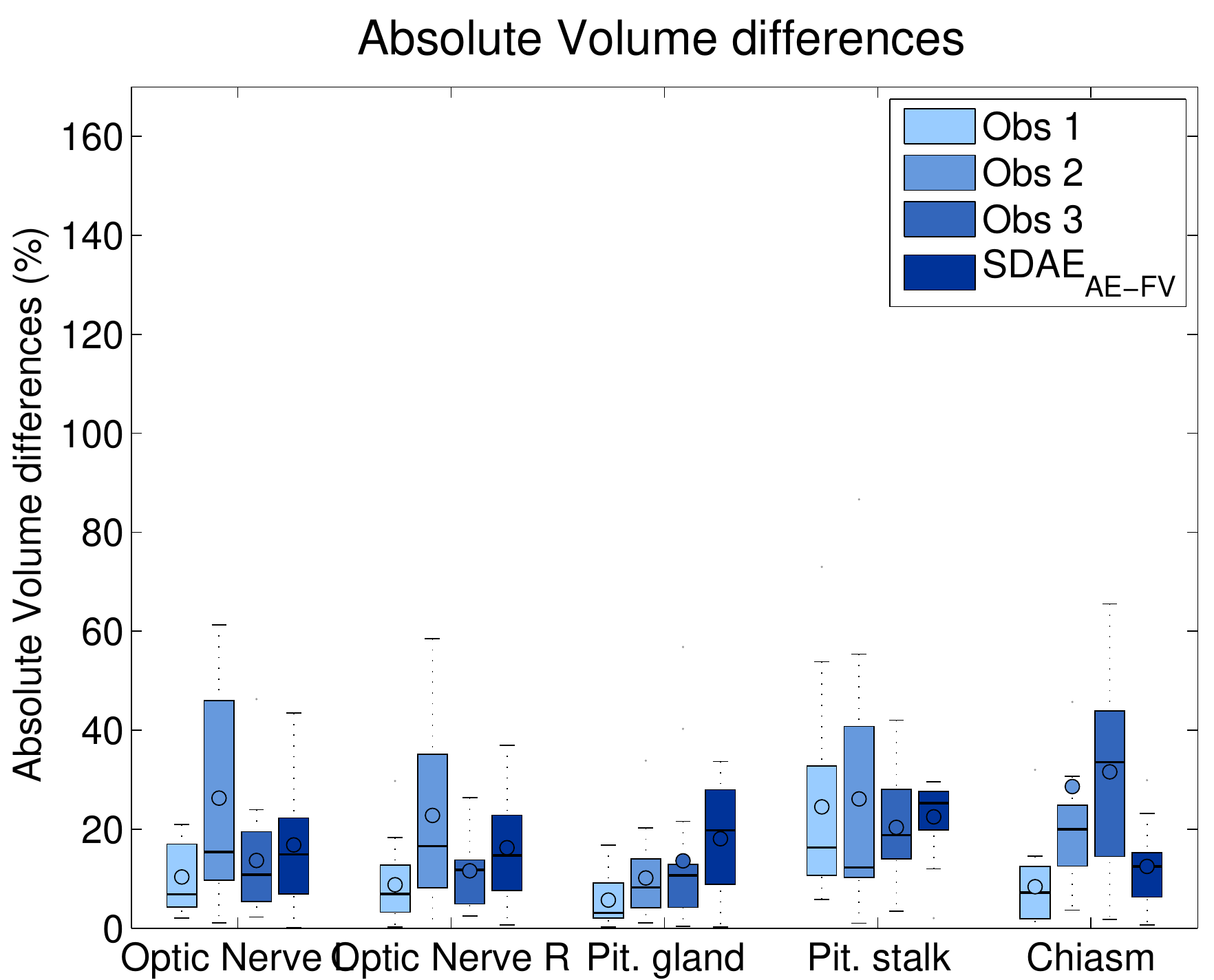}} 
\end{tabular}
\caption{Quantitative results of the performance of the six automatic configurations (\textit{left}) and the three observers (\textit{right}) in regards of volume similarities for the three OARs.}
\label{fig:Results}
\end{figure}

Since support vector machines (SVM) \cite{abe2010support} has proven to be a state-of-the-art classifier, we use it in this work for comparison purposes. Several configurations were evaluated, which changes comprise: i) the use of either SVM or SDAE for classification and ii) the use of one of the features sets described in table \ref{tab:features}. Accordingly, the first configuration will always be composed by classical features, being referred to as SVM$_1$ or SDAE$_1$, depending on whether it employs SVM or SDAE as classifier. Depending on the features set, configurations will be referred to as SDAE$_n$, where $n$ denotes the features group used, i.e \textit{augmented, textural or AE-FV}. Finally, SVM will be employed with the proposed features set leading to the SVM$_{AE-FV}$ configuration. Figure \ref{fig:Results} presents quantitative results of the performance of the six configurations in regards of volume similarities (\textit{left}), as well as results from manual raters when compared with the generated ground truth (\textit{right}). A detailed analysis of the results is presented below.

\paragraph{Dice Similarity Coefficient (DSC)} Distributions of DSC for the six configurations and for the three manual contours are plotted in figure \ref{fig:Results} (a) and (b), respectively. Box plots are grouped for each OAR. Inside each group, results for SVM references, and the several SDAE settings are displayed. Among all configurations, SVM based classifiers presented the lowest overall mean DSC values. Concerning the SDAE settings, the system that included our proposed features, SDAE$_{AE-FV}$, achieved the highest mean DSC value over all the OARs. Analyzing each structure separately, we can observe that again, mean DSC values from SVM configurations were among the lowest ones. In this setting, adding the set of proposed features generally improved the mean DSC. Nevertheless, it often remained below mean values achieved by SDAE based classifiers. Regarding the impact of different features sets on deep architectures, the use of classical features produced segmentations with acceptable mean DSC across all the OARs. However, it did not improve any of the other three features groups. Introduction of either augmented or textural features improved the segmentation performance of the classifier, which is reflected on its mean DSC values. Last, the use of the proposed features set, i.e. AE-FV, achieved the highest mean DSC values across all the structures with values of 0.78 ($\pm$ 0.05), 0.80 ($\pm$ 0.06), 0.76 ($\pm$ 0.06), 0.77 ($\pm$ 0.08) and 0.83 ($\pm$ 0.06), for left and right optic nerve, pituitary gland, pituitary stalk and chiasm, respectively.

Regarding to manual delineations, it can be observed on figure \ref{fig:Results} (b) that mean DSC achieved by the proposed system is always between the highest and lowest values reported by manual segmentations when compared with the reference standard. 


\paragraph{Hausdorff distances (HD)} Figure \ref{fig:Results} (b) and (d) plot the distribution of HD across the OARs for all the automatic frameworks and manual raters, respectively. As in the case of DSC distributions, mean HD values over all the structures show that SVM based classifiers presented the worst results. While SVM$_{AE-FV}$ achieved an overall mean HD of 6.63 ($\pm$ 5.09) mm, mean value for our proposed SDAE setting was of 3.32 ($\pm$ 0.96) mm. Looking at each structure individually, it can be observed that including the set of proposed features into the SVM system decreased mean HD values with respect to the classical features set when segmenting both optic nerves. For the rest of the organs, however, inclusion of proposed features did not particularly improve HD values. Employing SDAE as classifier instead of SVM in a classical features setting decreased mean HD in most cases. Incorporation of either augmented or textural features in the SDAE based classifier improved HD values with respect to classical features. While in some organs mean HD values were lower for augmented features based classifiers, for some other organs textural features set achieved the lowest mean HD values. Nevertheless, the combination of both features sets into the AE-FV set led to the lowest mean HD values across all the structures.



Although minimum HD values were not decreased when employing the deep learning scheme (2.58-3.67 mm), they ranged inside the variability of the experts (1.78-4.47 mm) or very close to values obtained by manual delineation. Furthermore, variability of reported HD values was decreased by the proposed system for some organs in comparison to some observers. Such is the case in both optic nerves in relation with observer 2 and 3. Variability of HD in segmenting the left optic nerve by observer 2 and 3 was of 1.96 and 1.32. Respectively, HD variability of right optic nerve was 1.89 and 0.85. By employing the proposed system this variability decreased to 0.87 and 0.66 for the left and right optic nerve, respectively.

\paragraph{Relative volume differences (rVD)} Distributions of relative volume differences are plotted in figure \ref{fig:Results} (c) and (e). Schemes employing SVM as classifier presented the largest volume differences for all the OARs. Indeed, with exception of the pituitary stalk, mean rVD for SVM based system were double than those reported by SDAE settings, independently on the features set used. Taking results from each structure, it can be observed that by employing either augmented or textural features in SDAE settings did not reduce mean rVD with respect to classical features. However, the proposed AE-FV set achieved the lowest rVD among all the configurations.


Segmentations from observer 1 presented the lowest mean rVD among the four groups (11.55$\%$ $\pm$ 12.78) over all the OARs. Mean rVD over all the OARs for segmentations of observer 2 and 3 were 22.80$\%$ ($\pm$ 25.24) and 18.17$\%$ ($\pm$ 15.11), respectively. Isolating results by group and organ, segmentations from observer 1 achieved the lowest mean rVD values across all the OARs.  For both optic nerves and pituitary stalk, contours from observer 2 obtained the highest mean rVD values, whilst observer 3 produced segmentations with highest mean rVD values for the chiasm. Our method was ranked at last when segmenting the pituitary gland, with a mean rVD value of 18.09$\%$. 

%

\paragraph{Statistical analysis}  

An ANOVA statistical analysis pointed out that our proposed system presented results significantly different from any other group in regards to DSC (p $<$ 0.05). Concerning HD values, differences between our approach and SVM based classifiers were significant in all the OARs. The use of proposed features against classical features in SDAE settings also presented significant differences, when segmentating both optic nerves and chiasm, with p-values of 0.0377, 0.0057 and 0.0165, respectively. Regarding rVD, results achieved by our system were significantly different than results from SVM settings in all the organs, with exception of the pituitary stalk (p = 0.7652). On the other hand, the impact of adding proposed features into the deep learning scheme was statistically significant only when segmenting the pituitary stalk and chiasm (p=0.0394 and p=0.0068). Compared to manual delineations, results achieved by the proposed approach did not present significant differences in most cases. Furthermore, in cases where differences where significantly different (p $<$ 0.05), our method outperformed the manual rater that presented those differences.

\paragraph{Sensitivity and specificity} 

Sensitivity and specificity across OARs for all configurations are reported in table \ref{table:SensSpecTableOARs3}. In general, SDAE based classifiers achieved the highest sensitivity values, whereas SVM settings obtained the highest specificity rates. Mean sensitivity values for both SVM configurations commonly ranged between 60 and 70 $\%$, with exception of the pituitary stalk, where sensitivity was around 70$\%$ for SVM$_1$ and close to 80$\%$ for SVM$_{AE-FV}$. Employing the SDAE system with classical features improved sensitivity, leading to values close to 80$\%$ for all the organs with exception of the chiasm, which mean sensitivity value was 71.67$\%$. Adding any single of the investigated features set (SDAE$_{Augmented}$ or SDAE$_{Textural}$) typically increased sensitivity with respect to classical settings. At last, the proposed system achieved sensitivity values greater than 80$\%$ in all the structures. Contrariwise, any pattern was identified concerning the sensitivity results. Combination of higher sensitivity and specificity metrics obtained from the AE-FV based classifier indicated that the proposed system correctly identified more tissue voxels than the others settings did, and also was better at rejecting tissue voxels that were not related to the tissue class of interest.


\begin{table}[h!]
\scriptsize
\centering
\begin{tabular}{|l|l|c|c|}
\hline
\textbf{}  & \textbf{Configuration}                                                            & \textbf{Sensitivity} & \textbf{Specificity} \\ \hline
\textbf{Optic nerve (L)} & \begin{tabular}[c]{@{}l@{}}SVM$_1$\\ SVM$_{AE-FV}$\\ SDAE$_1$\\ SDAE$_{Augmented}$\\ SDAE$_{Textural}$\\ SDAE$_{AE-FV}$\end{tabular} 
&
\begin{tabular}[c]{@{}c@{}} 66.68 ($\pm$ 10.74)
\\ 67.46 ($\pm$ 5.69)
\\ 85.41 ($\pm$ 5.76)
\\ 79.18 ($\pm$ 4.01)
\\ 81.87 ($\pm$ 3.49)
\\ 82.23 ($\pm$ 3.71)
\end{tabular}   
&
\begin{tabular}[c]{@{}c@{}} 79.19 ($\pm$ 23.57)
\\  92.86 ($\pm$ 6.64)
\\  79.38 ($\pm$ 15.07)
\\  90.44 ($\pm$ 7.27)
\\  89.34 ($\pm$ 7.65)
\\  91.02 ($\pm$ 7.31)
\end{tabular} 
\\ \hline
\textbf{Optic nerve (R)} & \begin{tabular}[c]{@{}l@{}}SVM$_1$\\ SVM$_{AE-FV}$\\ SDAE$_1$\\ SDAE$_{Augmented}$\\ SDAE$_{Textural}$\\ SDAE$_{AE-FV}$\end{tabular}                                                                                   &
\begin{tabular}[c]{@{}c@{}} 64.74 ($\pm$ 12.81)
\\  66.31 ($\pm$ 8.68)
\\  79.30 ($\pm$ 6.13)
\\  80.53 ($\pm$ 5.18)
\\  80.19 ($\pm$ 4.84)
\\  81.54 ($\pm$ 4.45)
\end{tabular}                   &   
\begin{tabular}[c]{@{}c@{}} 76.52 ($\pm$ 23.82)
\\   91.29 ($\pm$ 10.32)
\\   82.79 ($\pm$ 13.28)
\\   87.67 ($\pm$ 10.82)
\\   87.86 ($\pm$ 10.86)
\\   88.09 ($\pm$ 9.52)
\end{tabular}
\\ \hline
\textbf{Pituitary gland} &  \begin{tabular}[c]{@{}l@{}}SVM$_1$\\ SVM$_{AE-FV}$\\ SDAE$_1$\\ SDAE$_{Augmented}$\\ SDAE$_{Textural}$\\ SDAE$_{AE-FV}$\end{tabular}
&\begin{tabular}[c]{@{}c@{}}62.31 ($\pm$ 15.18)
\\ 67.81 ($\pm$ 14.89)
\\ 80.85 ($\pm$ 9.69)
\\ 83.13 ($\pm$ 9.29)
\\ 82.24 ($\pm$ 10.05)
\\ 84.22 ($\pm$ 7.94)\end{tabular}
&
\begin{tabular}[c]{@{}c@{}} 94.84 ($\pm$ 6.52)
\\ 88.51 ($\pm$ 10.62)
\\ 80.86 ($\pm$ 14.32)
\\ 79.85 ($\pm$ 19.35)
\\ 81.07 ($\pm$ 13.79)
\\ 82.69 ($\pm$ 15.09)
\end{tabular}                    
\\ \hline
\textbf{Pituitary stalk} & \begin{tabular}[c]{@{}l@{}}SVM$_1$\\ SVM$_{AE-FV}$\\ SDAE$_1$\\ SDAE$_{Augmented}$\\ SDAE$_{Textural}$\\ SDAE$_{AE-FV}$\end{tabular}                                                                                   &
\begin{tabular}[c]{@{}c@{}}70.33 ($\pm$ 6.94)
\\ 80.78 ($\pm$ 7.76)
\\ 79.19 ($\pm$ 8.02)
\\ 81.66 ($\pm$ 6.47)
\\ 79.62 ($\pm$ 8.17)
\\ 82.28 ($\pm$ 7.53)
\end{tabular}
&
\begin{tabular}[c]{@{}c@{}} 84.42 ($\pm$ 10.62)
\\ 77.61 ($\pm$ 14.54)
\\ 76.52 ($\pm$ 17.42)
\\ 77.29 ($\pm$ 14.28)
\\ 77.98 ($\pm$ 17.19)
\\ 73.14 ($\pm$ 16.86)
\end{tabular}                    
\\ \hline
\textbf{Chiasm} &  \begin{tabular}[c]{@{}l@{}}SVM$_1$\\ SVM$_{AE-FV}$\\ SDAE$_1$\\ SDAE$_{Augmented}$\\ SDAE$_{Textural}$\\ SDAE$_{AE-FV}$\end{tabular}                                                                                  &
\begin{tabular}[c]{@{}c@{}}65.09 ($\pm$ 7.78)
\\ 69.74 ($\pm$ 11.39)
\\ 71.67 ($\pm$ 12.07)
\\ 83.93 ($\pm$ 5.16)
\\ 84.32 ($\pm$ 7.40)
\\ 83.94 ($\pm$ 4.34)
\end{tabular}
&
\begin{tabular}[c]{@{}c@{}} 94.37 ($\pm$ 7.88)
\\ 88.43 ($\pm$ 10.57)
\\ 89.84 ($\pm$ 15.23)
\\ 86.64 ($\pm$ 9.69)
\\ 82.42 ($\pm$ 17.78)
\\ 86.11 ($\pm$ 9.71)
\end{tabular}                   \\ \hline
\end{tabular}
\caption{Sensitivity and specificity mean values for the six automatic configurations across the OARs.}
\label{table:SensSpecTableOARs3}
\end{table}

\begin{figure}[h!]
\centering
\begin{tabular}{cc}
\subfloat[Left optic nerve]{\includegraphics[width=0.45\linewidth]{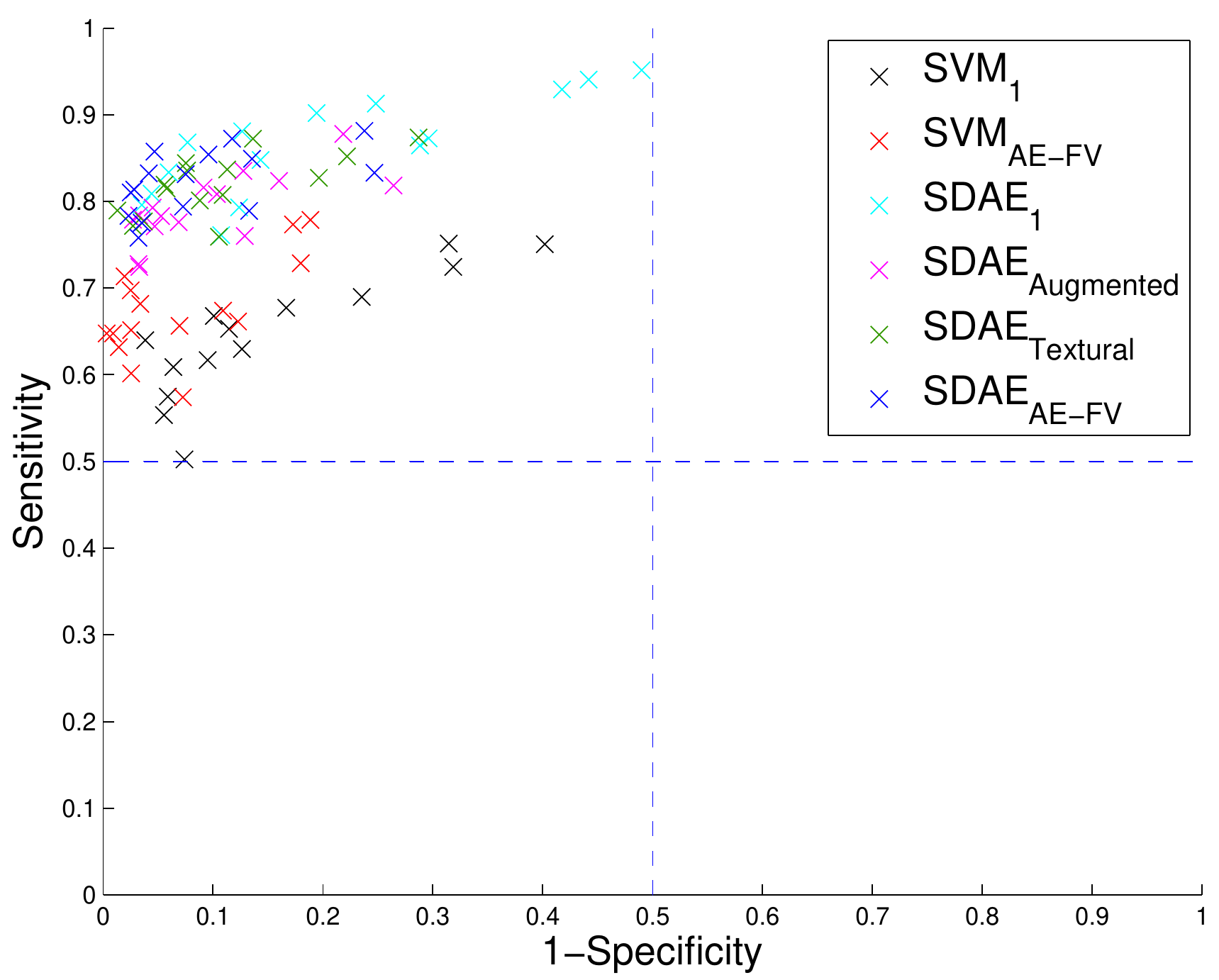}} & 
\subfloat[Right optic nerve]{\includegraphics[width=0.45\linewidth]{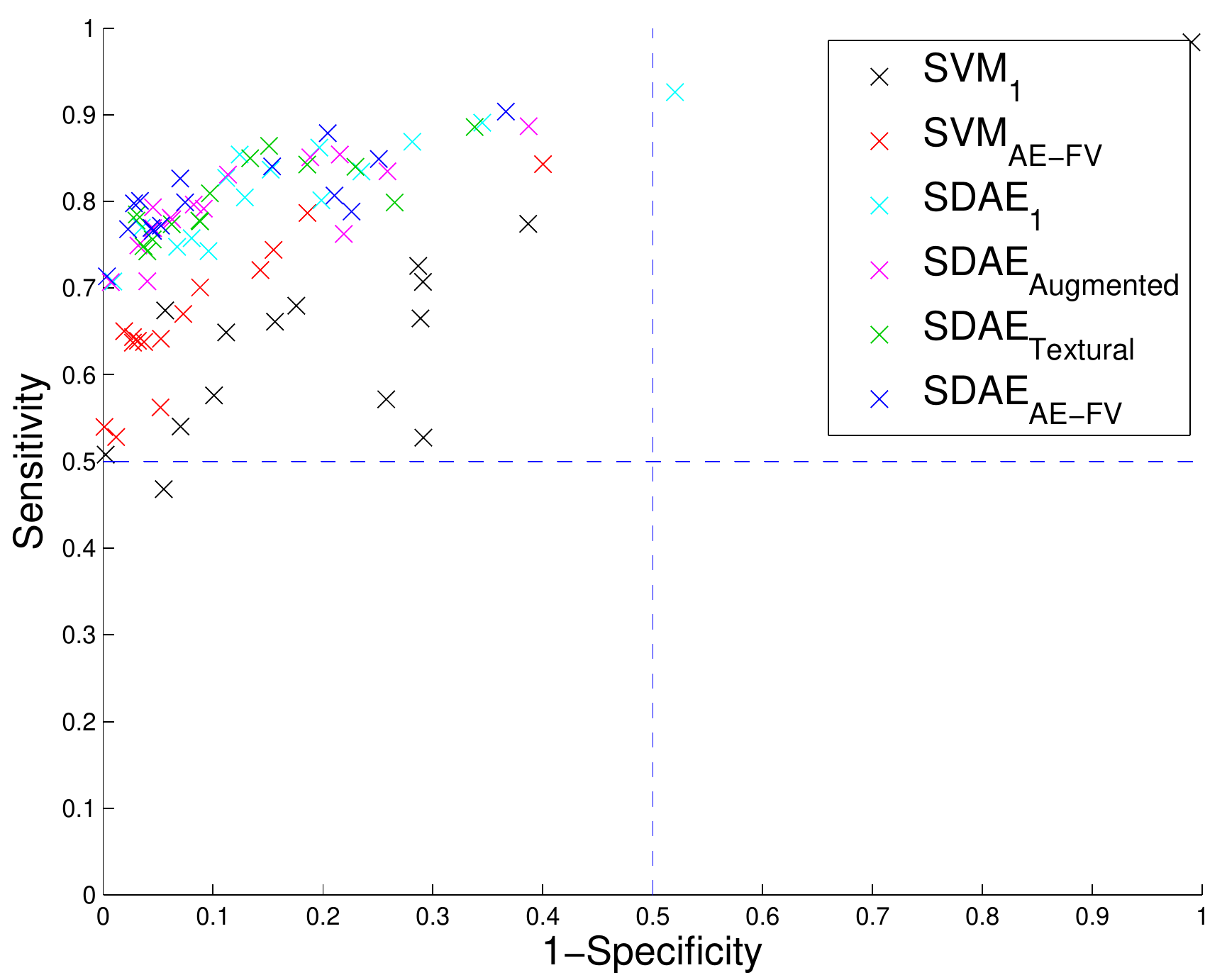}} \\ 
\subfloat[Pituitary gland]{\includegraphics[width=0.45\linewidth]{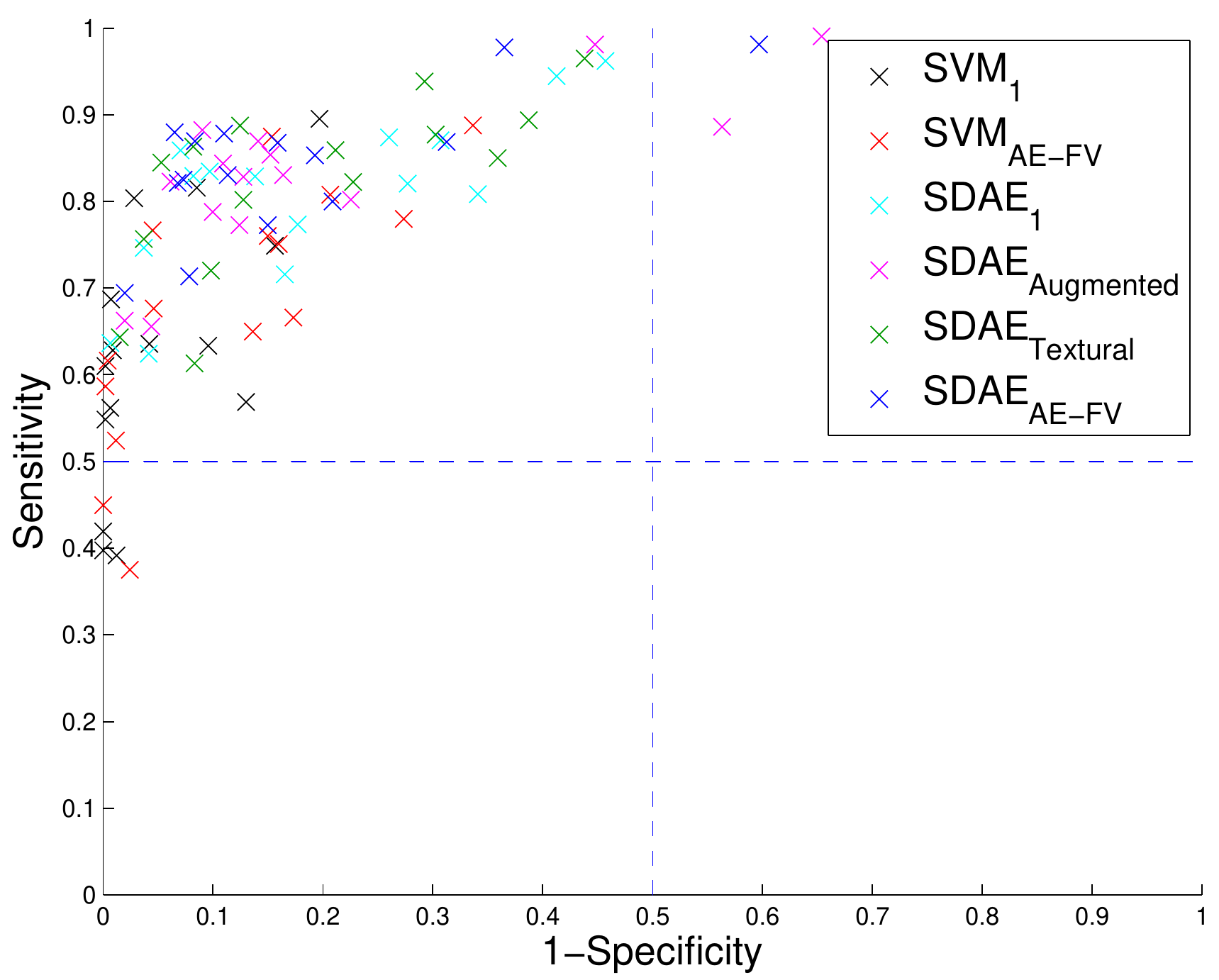}} &
\subfloat[Pituitary stalk]{\includegraphics[width=0.45\linewidth]{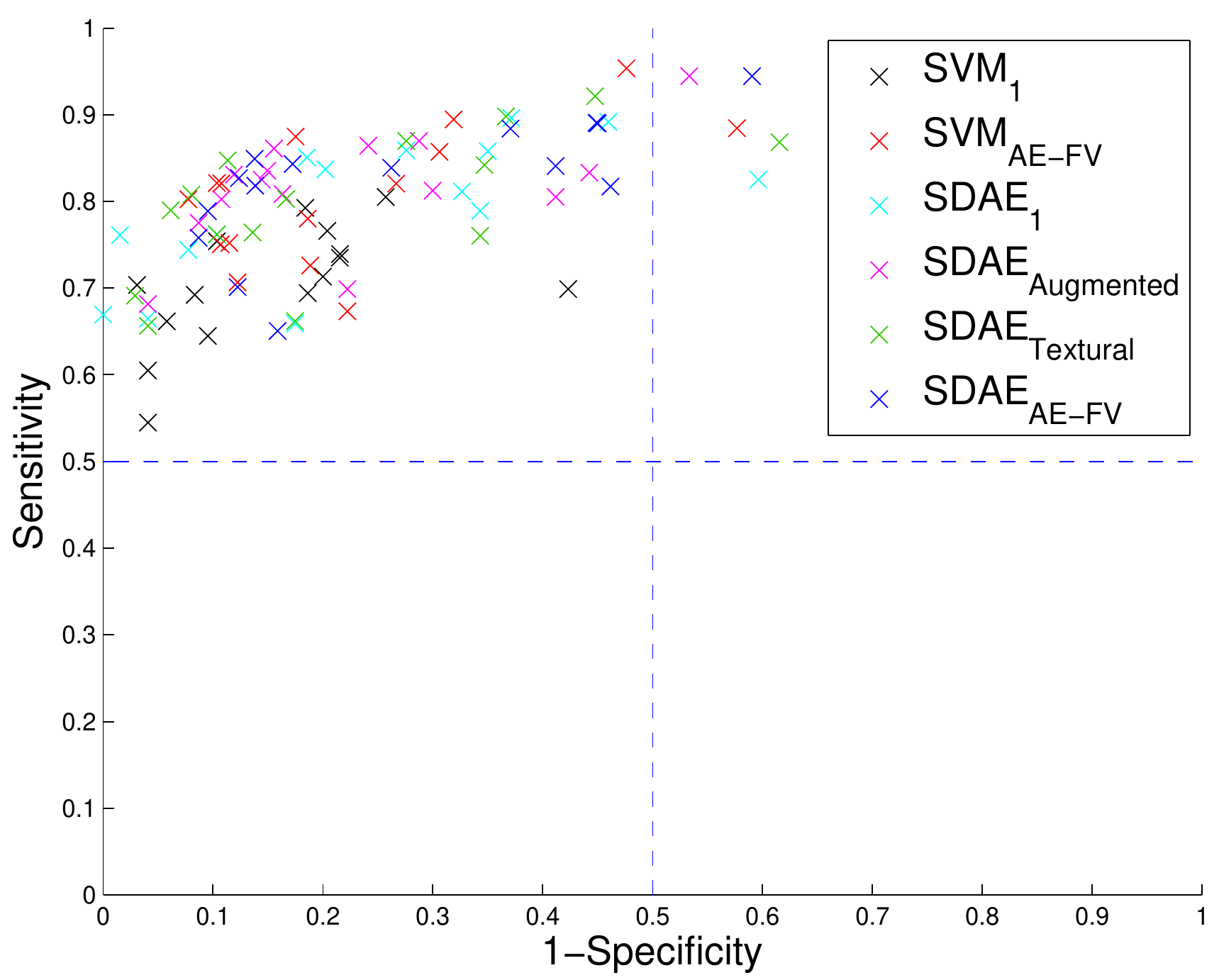}} \\ 
\subfloat[Optic chiasm]{\includegraphics[width=0.45\linewidth]{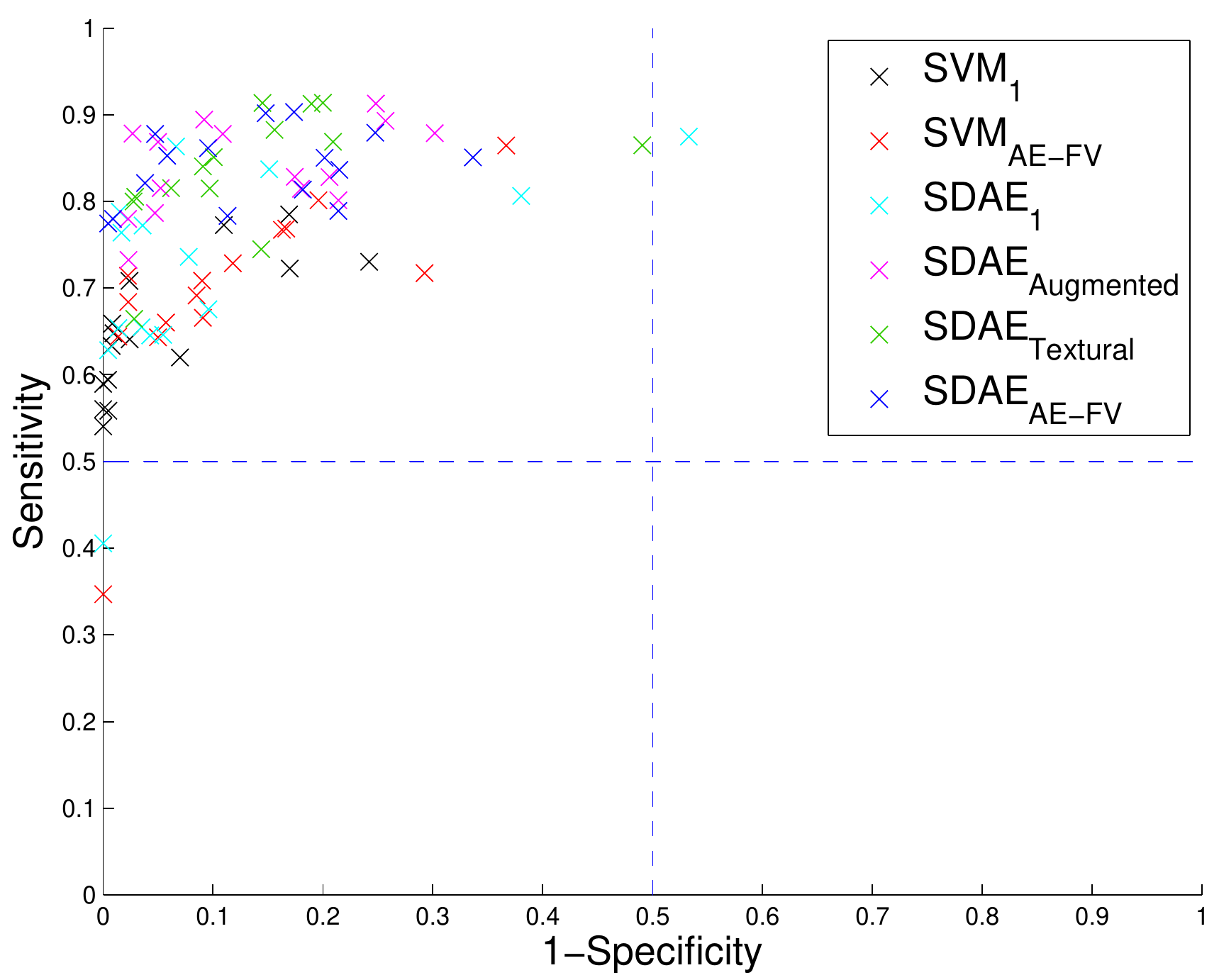}}
\end{tabular}
\caption{ROC sub-division analysis for the six automatic approaches for organs of group A.}
\label{fig:ROCArea_GroupB}
\end{figure}

The sub-division scheme proposed by \cite{andrews1985benefit} is applied to analyze the ROC of all the automatic settings (Figure \ref{fig:ROCArea_GroupB}). Each cross represents the correspondence between sensitivity and (1 - specificity) of a single patient and its color indicates the setting employed. First, it can be observed that for the six configurations nearly all results lie on the left-top sub-space, which indicates contours would be considered acceptable for RTP. Nevertheless, there are cases which should be taken into consideration. For example, some contours are inside the "high risk" area when segmenting pituitary gland and stalk, meaning that the OAR may be spared but the PTV not covered. In addition, although contours provided by both SVM configurations lie inside the "acceptable" area, they dangerously surround the "poor" region, where the OARs are not spared. 


\paragraph{Segmentation time.} Segmentation time is divided into two steps: features extraction and segmentation or classification. While features extraction was common for each features set and took between 1-4 seconds for an entire volume, classification depended on both classifier and features set employed.  Table \ref{table:segTimes} presents mean segmentation times for first and last features sets for both SVM and SDAE classifiers. Mean times for SVM based systems ranged from few seconds, in small structures, to one or several minutes in large structures or structures presenting large shape variations. The use of proposed features into the classifiers increased segmentation times, which is normal if we take into consideration that the proposed features set was composed by a larger number of features. SDAE based classification schemes achieved segmentations in less than a second for all the OARs.


\begin{table}[h!]
\centering
\scriptsize
\begin{tabular}{|l|c|c|c|c|}
\hline
\multicolumn{5}{|c|}{\textbf{Segmentation time (seconds)}}                    \\ \hline
& \textbf{SVM$_1$} & \textbf{SVM$_{Last}$} & \textbf{SDAE$_1$} & \textbf{SDAE$_{Last}$} \\ \hline
\textbf{Optic nerve (L)} & 173.4234 ($\pm$ 5.4534)  & 221.3296 ($\pm$ 6.7034)  & 0.1915 ($\pm$ 0.0124)  &   0.2628 ($\pm$ 0.0172)                  \\ \hline
\textbf{Optic nerve (R)} & 167.7524 ($\pm$ 6.7484)   & 214.4560 ($\pm$ 9.3614) & 0.1726 ($\pm$ 0.0091) & 0.2517 ($\pm$ 0.0194)  \\ \hline
\textbf{Pituitary gland} & 15.5368 ($\pm$ 0.7802) & 19.3440 ($\pm$ 0.8235) & 0.0536 ($\pm$ 0.0066) & 0.0748 ($\pm$ 0.0065) \\ \hline
\textbf{Pituitary stalk} & 3.0150 ($\pm$ 0.1485) & 4.1328 ($\pm$ 0.3899)  & 0.0146 ($\pm$ 0.0018) &   0.0262 ($\pm$ 0.0027)                \\ \hline
\textbf{Chiasm} & 5.2022 ($\pm$ 0.3214)  & 5.8751 ($\pm$ 0.5424)  & 0.0628 ($\pm$ 0.0065) & 0.1315 ($\pm$ 0.0124)          \\ \hline
\end{tabular}
\caption{Segmentation times.}
\label{table:segTimes}
\end{table}

\begin{figure}[h!]
\centering
\begin{tabular}{ccc}
\subfloat{\includegraphics[width=0.25\linewidth]{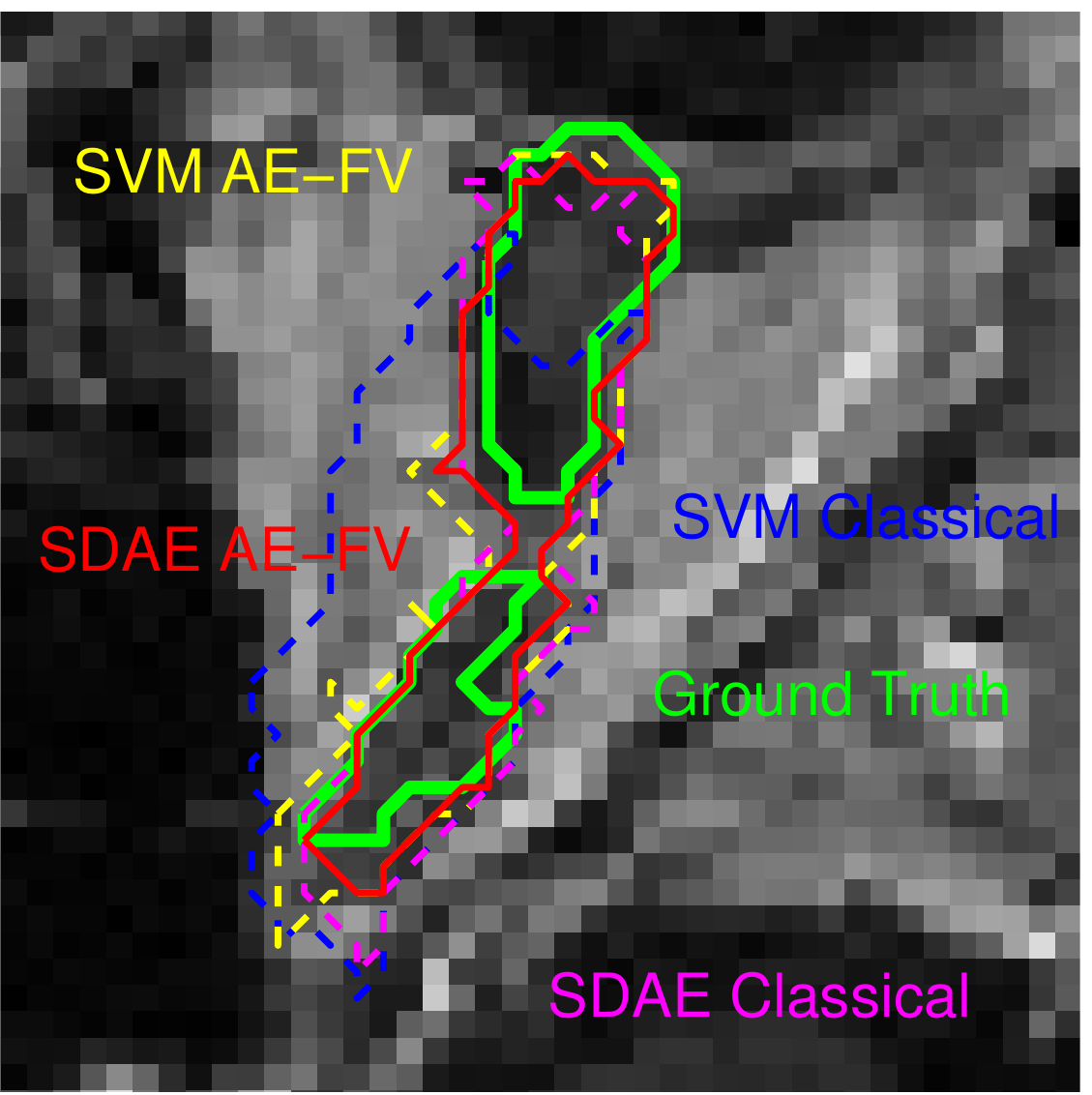}} 
\subfloat{\includegraphics[width=0.25\linewidth]{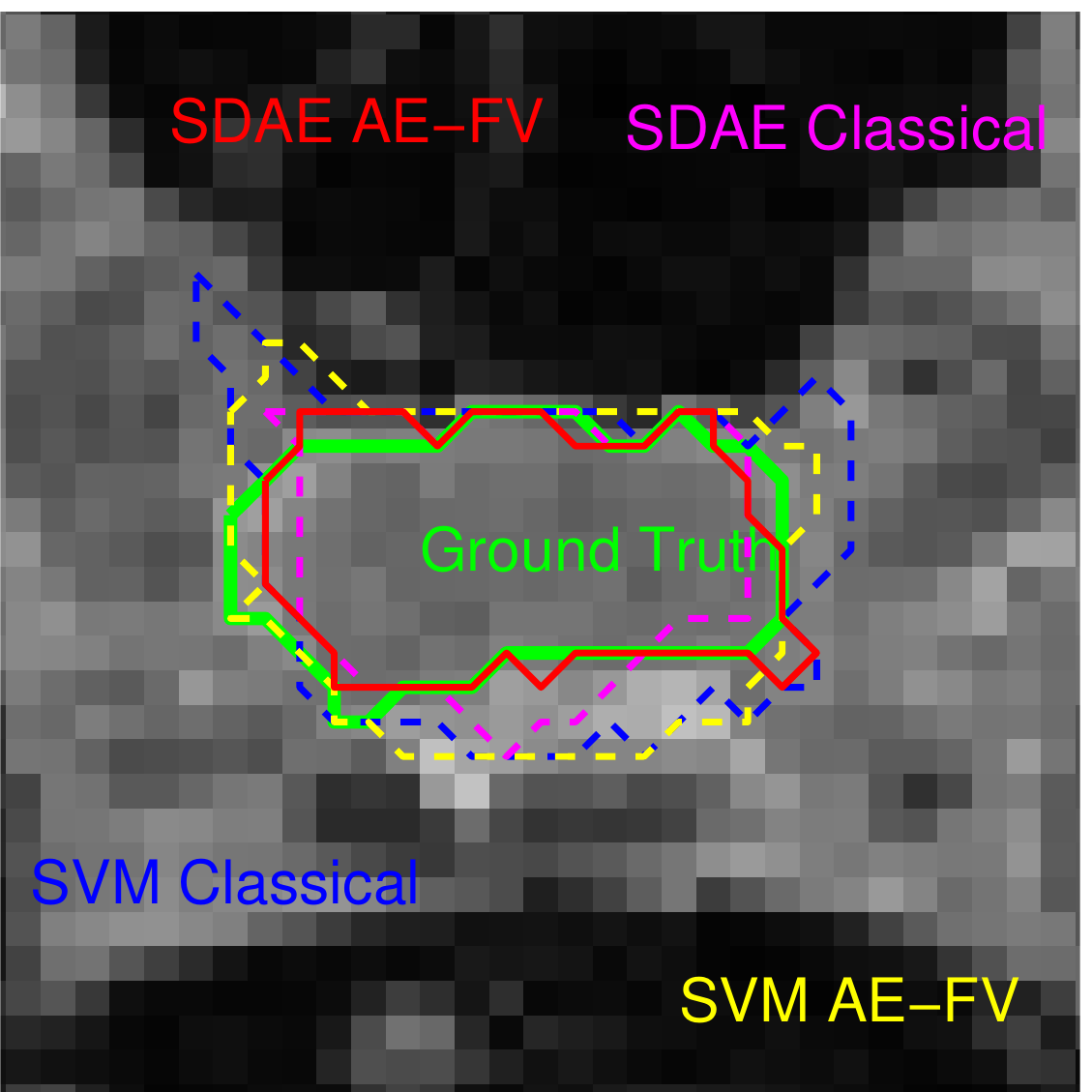}} 
\subfloat{\includegraphics[width=0.25\linewidth]{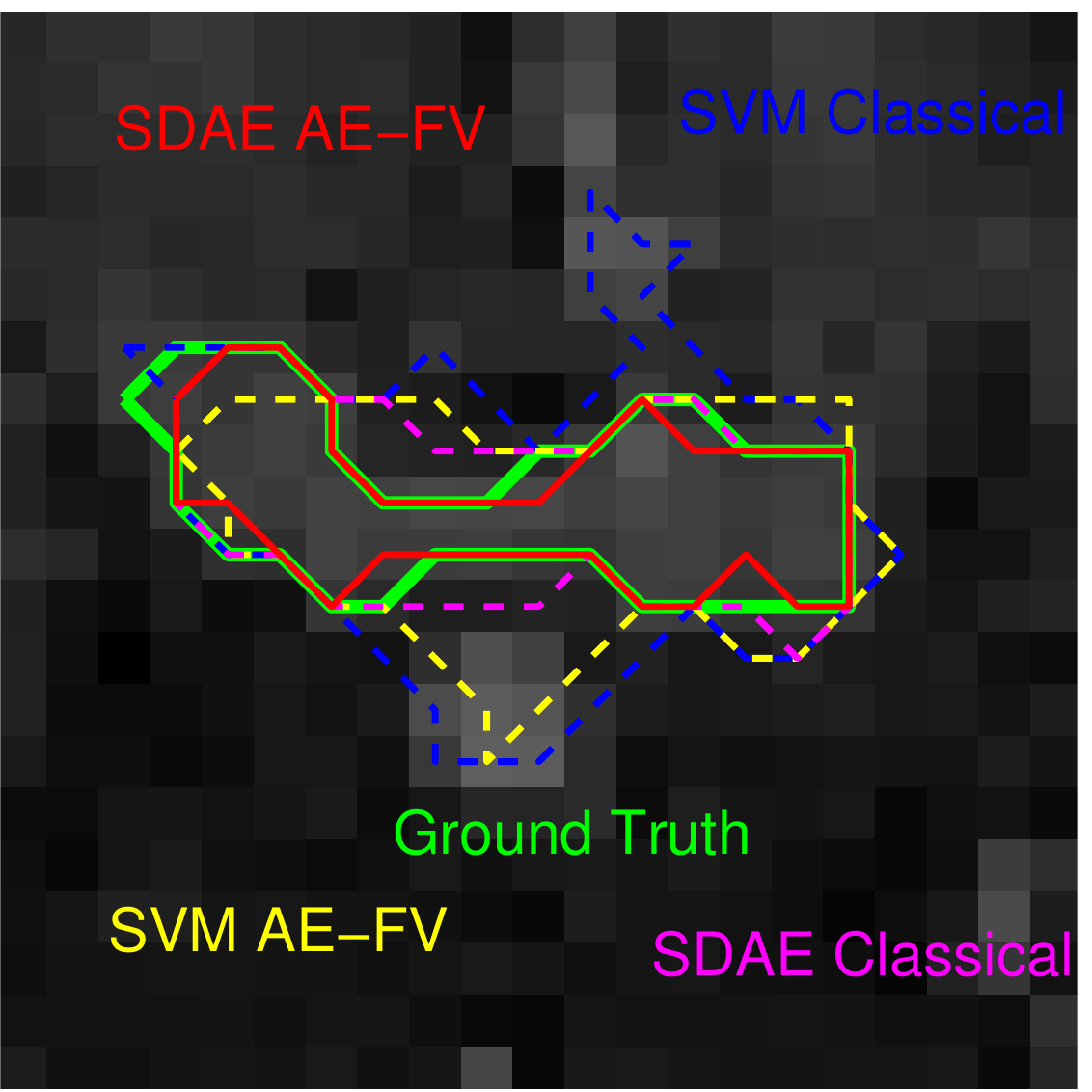}} \\ [-2ex]
\subfloat{\includegraphics[width=0.25\linewidth]{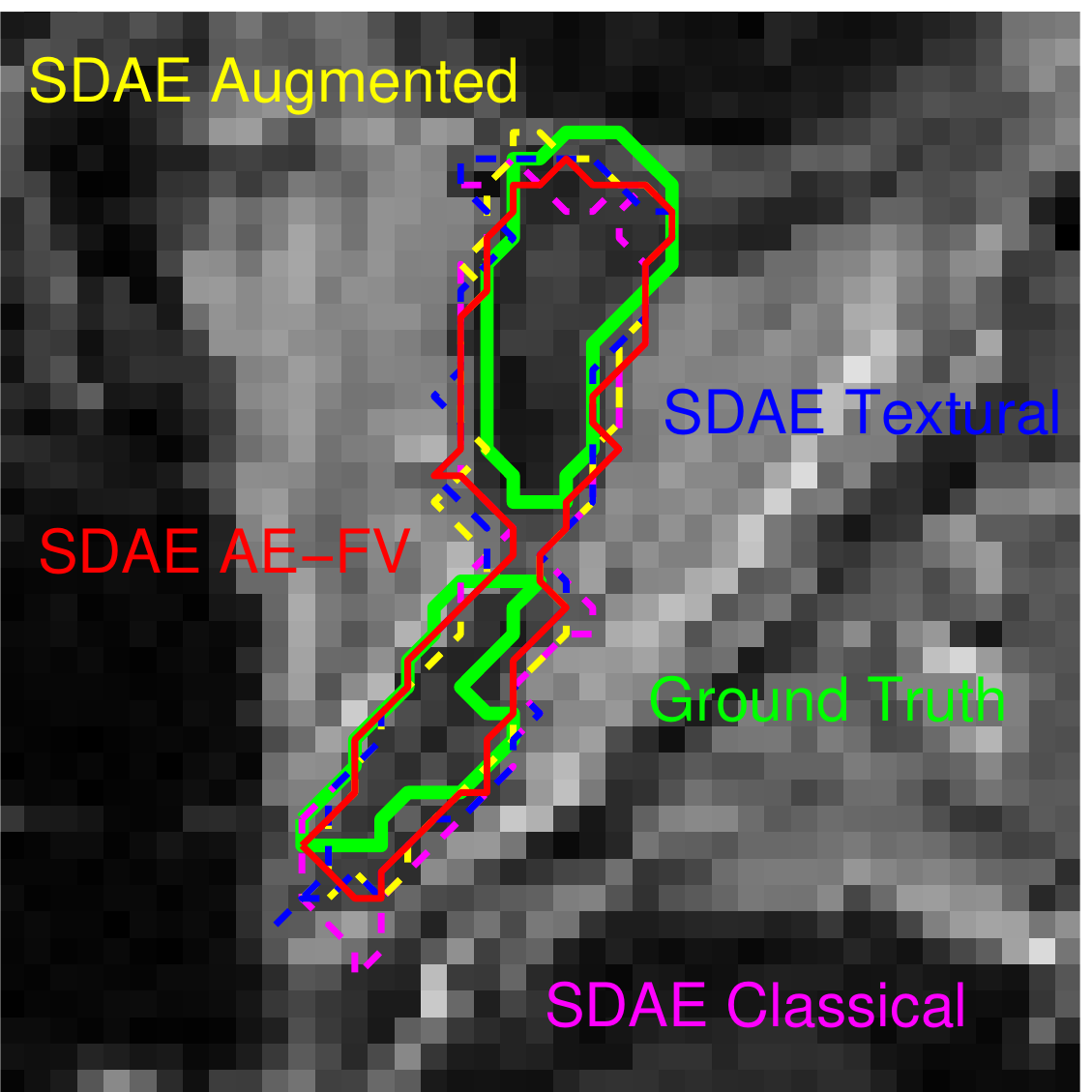}} 
\subfloat{\includegraphics[width=0.25\linewidth]{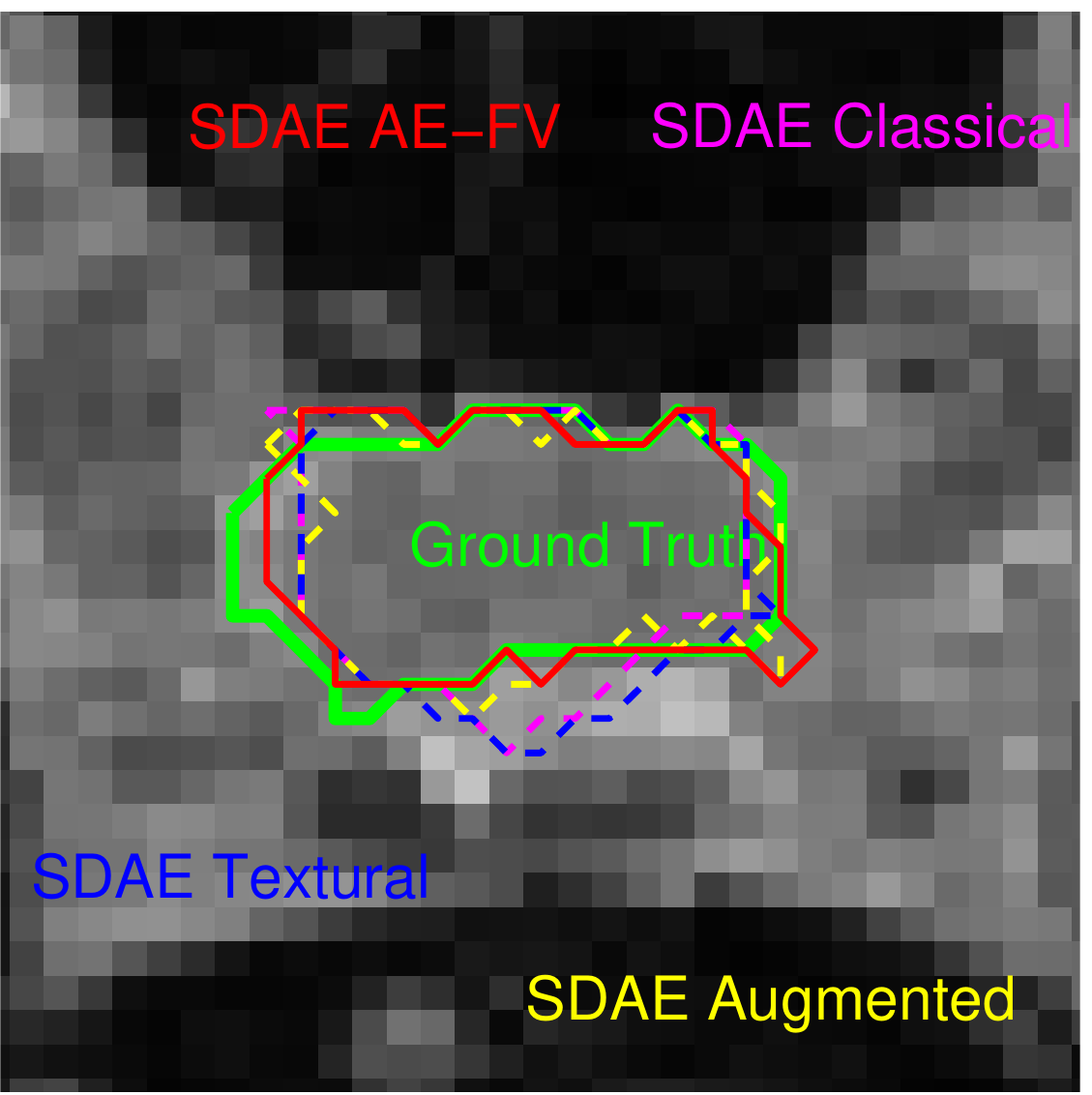}} 
\subfloat{\includegraphics[width=0.25\linewidth]{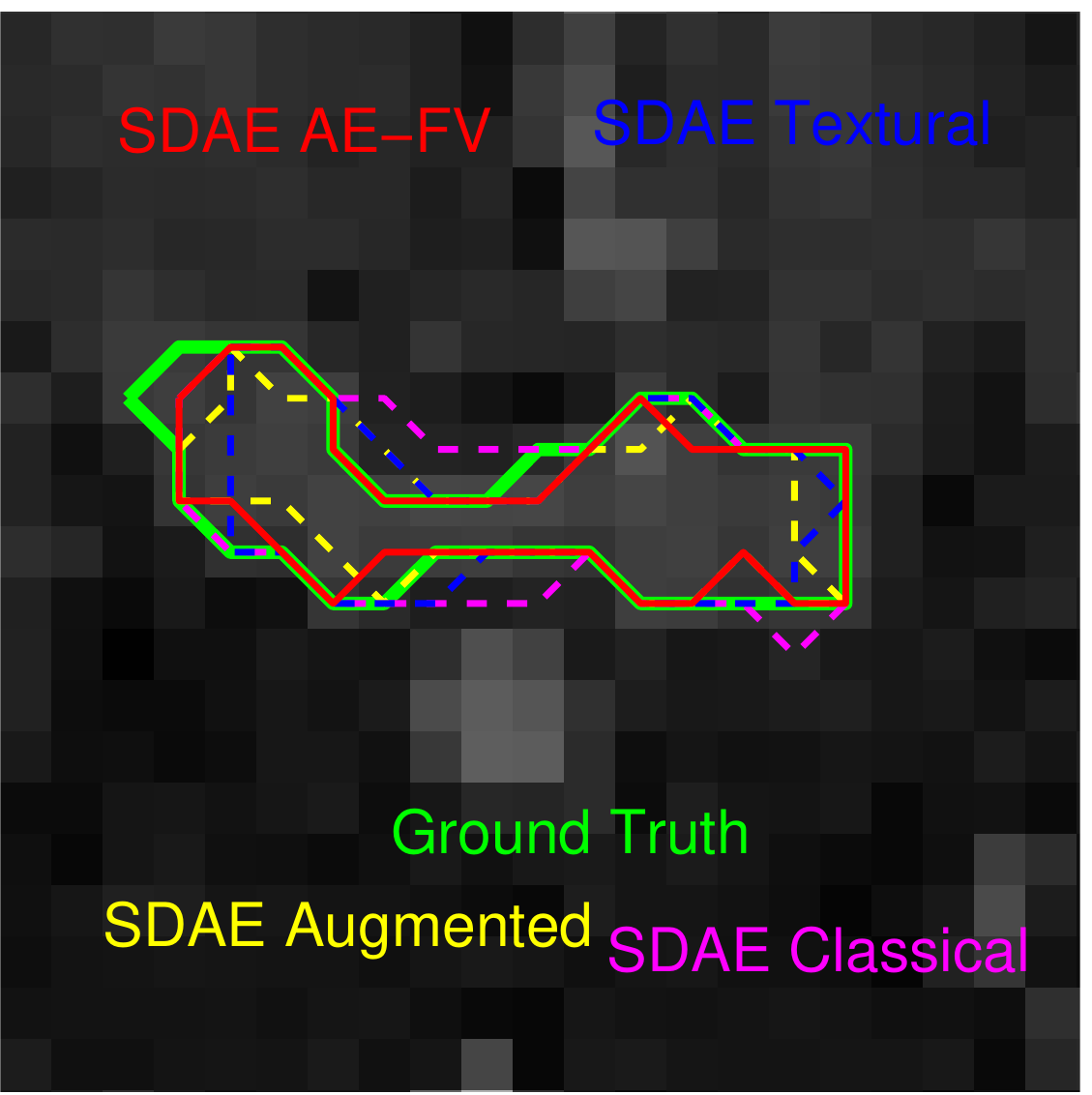}} 
\end{tabular}
\caption{Segmentation results produced by the proposed classification system when segmenting the right optic nerve (\textit{left}), pituitary gland (\textit{middle}) and chiasm (\textit{right}), and comparison with the other automatic configurations.}
\label{fig:VisualresultsAutoB}
\end{figure}

Figure \ref{fig:VisualresultsAutoB} displays the automatic contours generated by the evaluated configurations. To investigate the effect on the segmentation of employing different classifiers, segmentations from SVM and SDAE configurations are presented on the top-row. Visual results show that SVM based classifiers provided much larger contours than the ground truth. This was particularly noticeable in the contours from SVM with classical features. In the case of the chiasm, for example, SVM configurations were not capable of distinguish between chiasm and pituitary stalk. Contrary, classifiers based on SDAE correctly classified the chiasm avoiding the neighboring region of the pituitary stalk. Comparison of the impact on the segmentation performance when adding the different features sets on the SDAE settings can be seen in the bottom-row. Including either augmented or textural features into the classification system typically improved segmentations with respect to classical features. Nevertheless, combining all features into the AE-FV set achieved the best contours among the SDAE frameworks.

\section{Discussion}
\label{sec:discussion}
A deep learning-based classification scheme created by stacking denoising auto-encoders has been proposed in this work to segment organs at risk in the optic region in brain cancer. One of the main contributions of this paper is the incorporation of contextual, first order texture statistics and spectral features into the features vector as input of the deep network. Segmentation accuracy is improved by including all this information into the classification process. This work is not pioneering on the use of a stack of DAE (SDAE) to segment OARs in brain cancer. In \cite{DolzCMIG}, a similar approach was proposed to segment the brainstem. Nevertheless, this work presents some differences. First, we propose an approach that is not tailored to only one structure. Furthermore, structures segmented in the presented experiment are more complex to segment. And third, the features set employed in \cite{DolzCMIG} is extended. On the other hand, clinical evaluation of our automatic system involving manual segmentations from several experts was also assessed in our experiment.

We have explored in this paper whether it is plausible to use hand-crafted features to train a deep learning network to segment small OARs in brain cancer. We have noticed that deep learning is recently becoming quite popular in the medical domain. Although its application to medical imaging is being explored, features sets employed in most of ongoing works are very different from the set proposed in this work. To demonstrate that the union of contextual and textural features into an enhanced features array can improve the performance we investigated the impact on the segmentation of several sets of features. By adding any of these types of features to the classical features array, an improvement was already noticeable. Across the experiment we noticed that, while in some patients the use of augmented features achieved better results, in other patients the result was improved when using textural features instead. However, when combining both of them results were more homogeneous, which can be also observed in the standard deviation of results.

Although several attempts to segment some of these small structures have been presented, unsatisfactory results have been reported. Among the four OARs analyzed in the present study, the optic nerves and chiasm have received the most attention. In various evaluations performed in RTP context, \cite{dhaese,isambert,bekes2008geometrical,Deeley}, automatic segmentations were not sufficiently accurate to be usable in RTP. More recently, \cite{noble2011atlas} presented an atlas-based algorithm, which combined CT and MR images to segment optic nerves and chiasm, which achieved a mean DSC value just below 0.8 for both structures. Nevertheless, a computation time close to 20 minutes was reported. Although in terms of similarity the proposed approach is comparable to their work to segment the chiasm, important differences are two-fold. First, it does not require combination of image modalities. Second, segmentation time is largely faster than proposed approaches. In another recent study on RTP context \cite{Deeley}, manual and automated approaches were compared to segment brain structures in the presence of space-occupying lesions. To achieve the automation of the segmentation process, a registration-driven atlas-based algorithm was employed. A set comprising the brainstem, eyes, optic chiasm and optic nerves was evaluated. Main results disclosed in their evaluation showed that the analyzed automatic approach exhibited mean DSC values between 0.8-0.85 for larger structures. Contrary, DSC reported for smaller structures, i.e. optic chiasm and optic nerves, were of 0.4 and 0.5, respectively. Regarding others structures, only \cite{isambert} included the pituitary gland on their evaluation, with no success at all. Therefore and to the best of our knowledge, results suggest that the method proposed in this work is the most accurate, robust and fast method to date to accomplish automatic segmentation of optic nerves, optic chiasm, pituitary gland and pituitary stalk.

\begin{figure}[h!]
\centering
\begin{tabular}{cccc}
\subfloat{\includegraphics[width=0.21\linewidth]{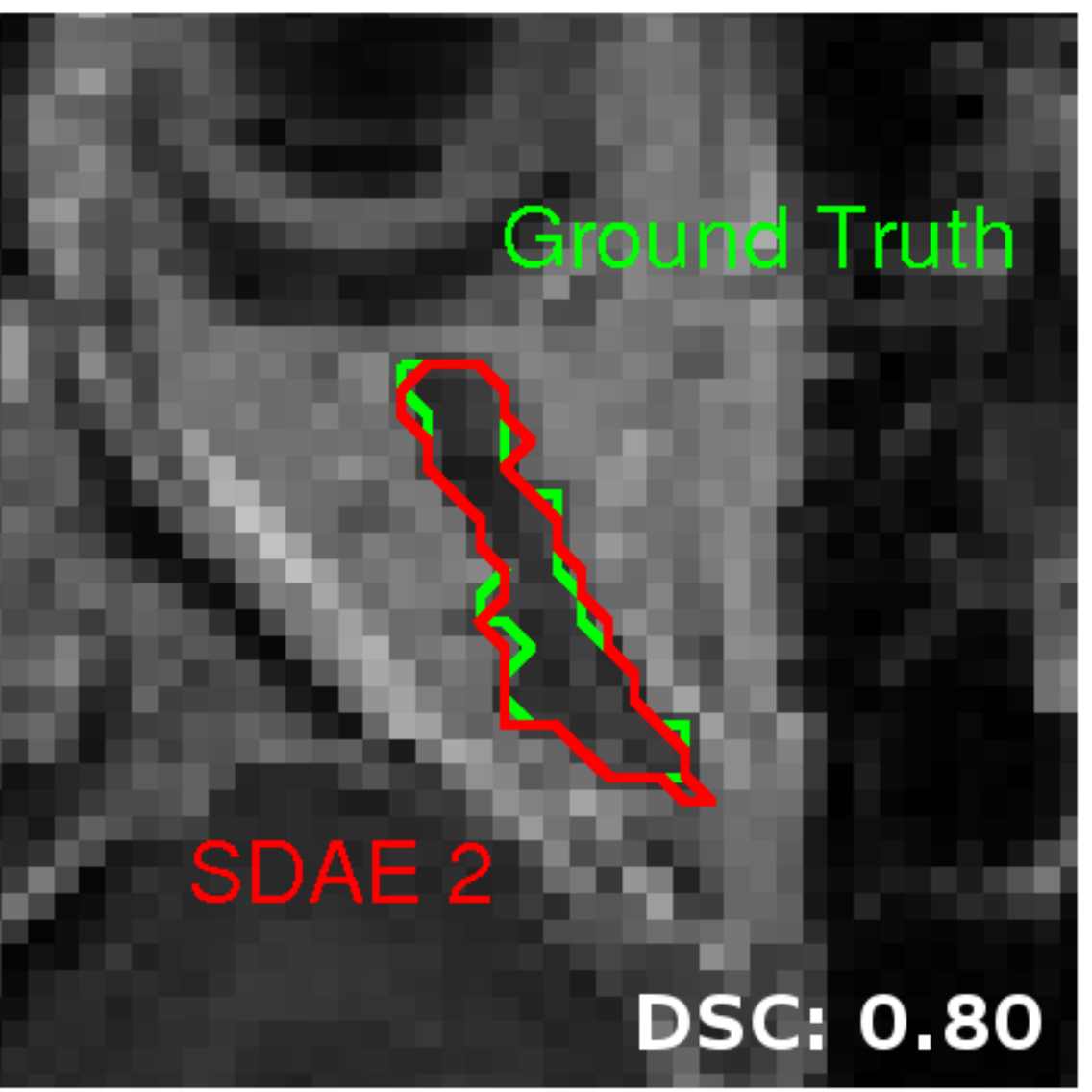}} \hspace{0.01em}
\subfloat{\includegraphics[width=0.21\linewidth]{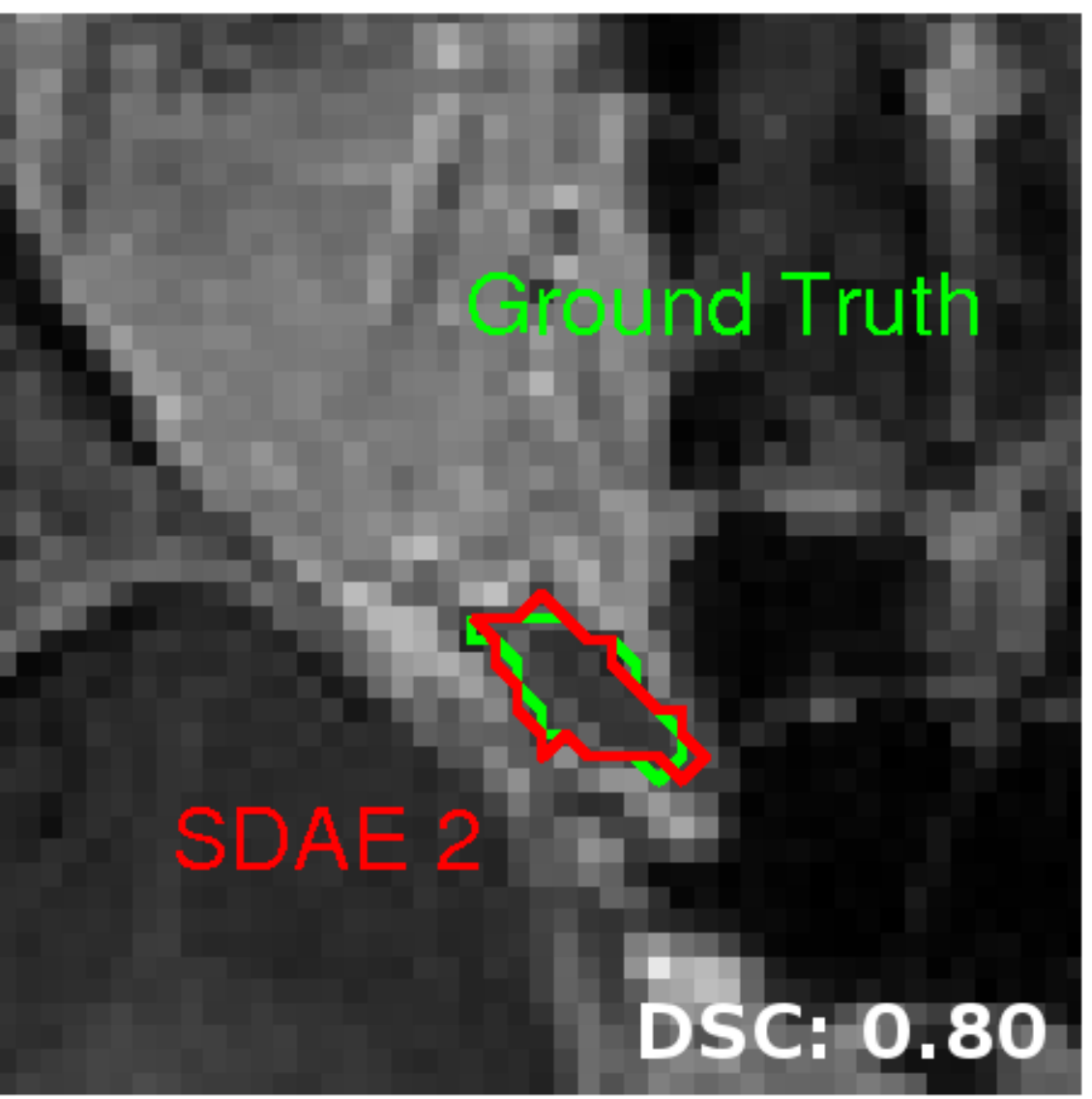}} \hspace{0.01em}
\subfloat{\includegraphics[width=0.21\linewidth]{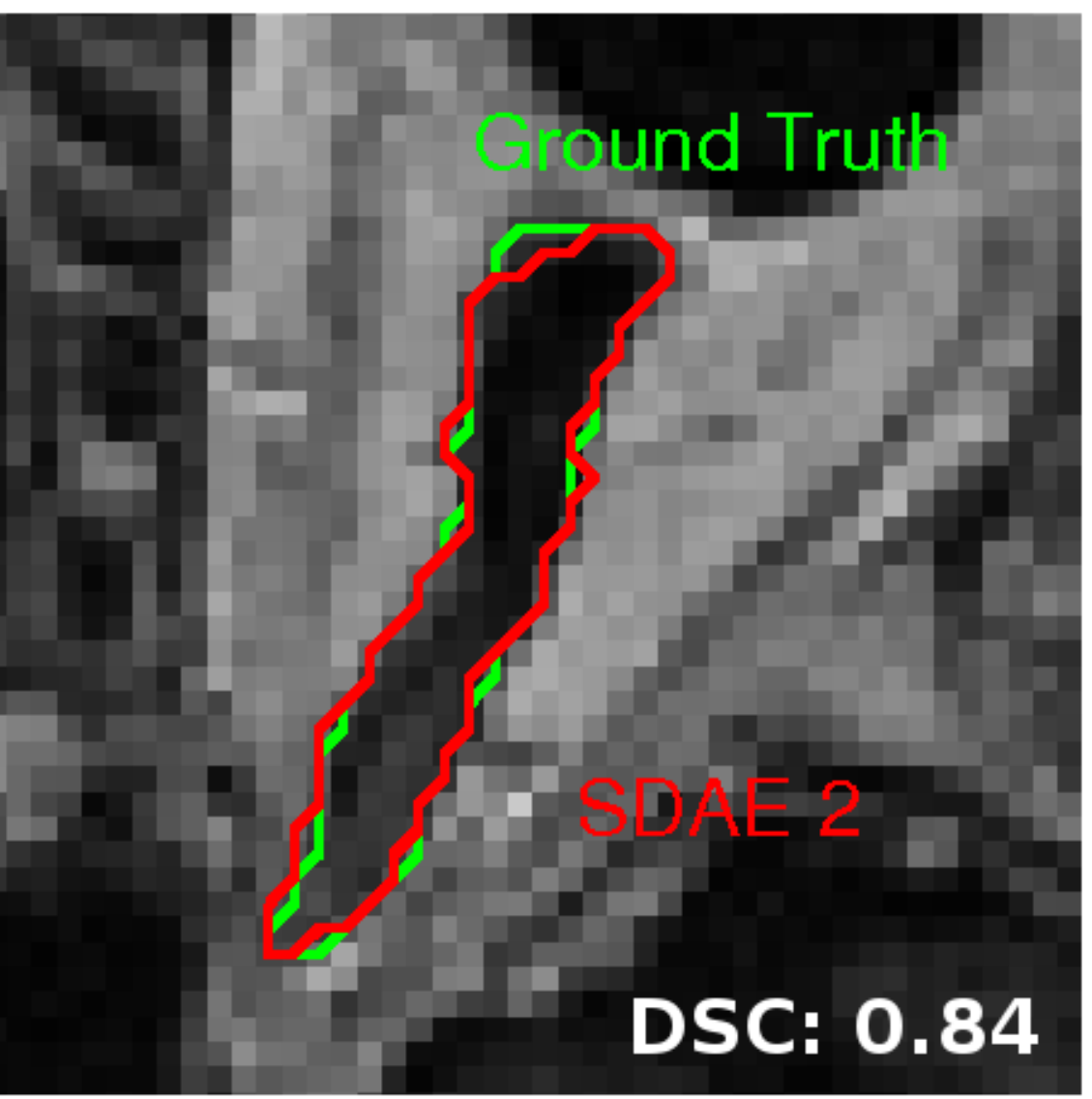}} \hspace{0.01em}
\subfloat{\includegraphics[width=0.21\linewidth]{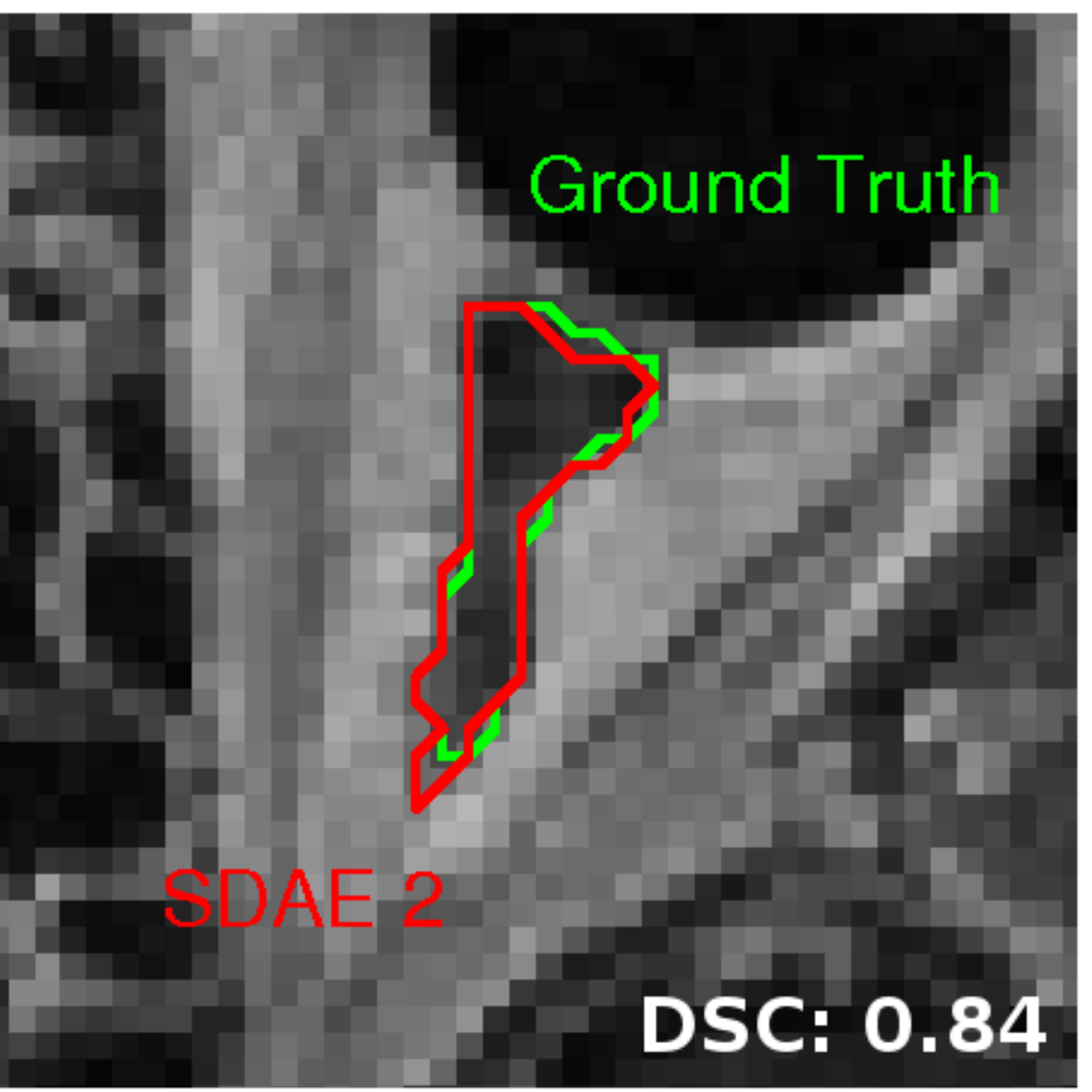}}  \\ [-2ex]
\subfloat{\includegraphics[width=0.21\linewidth]{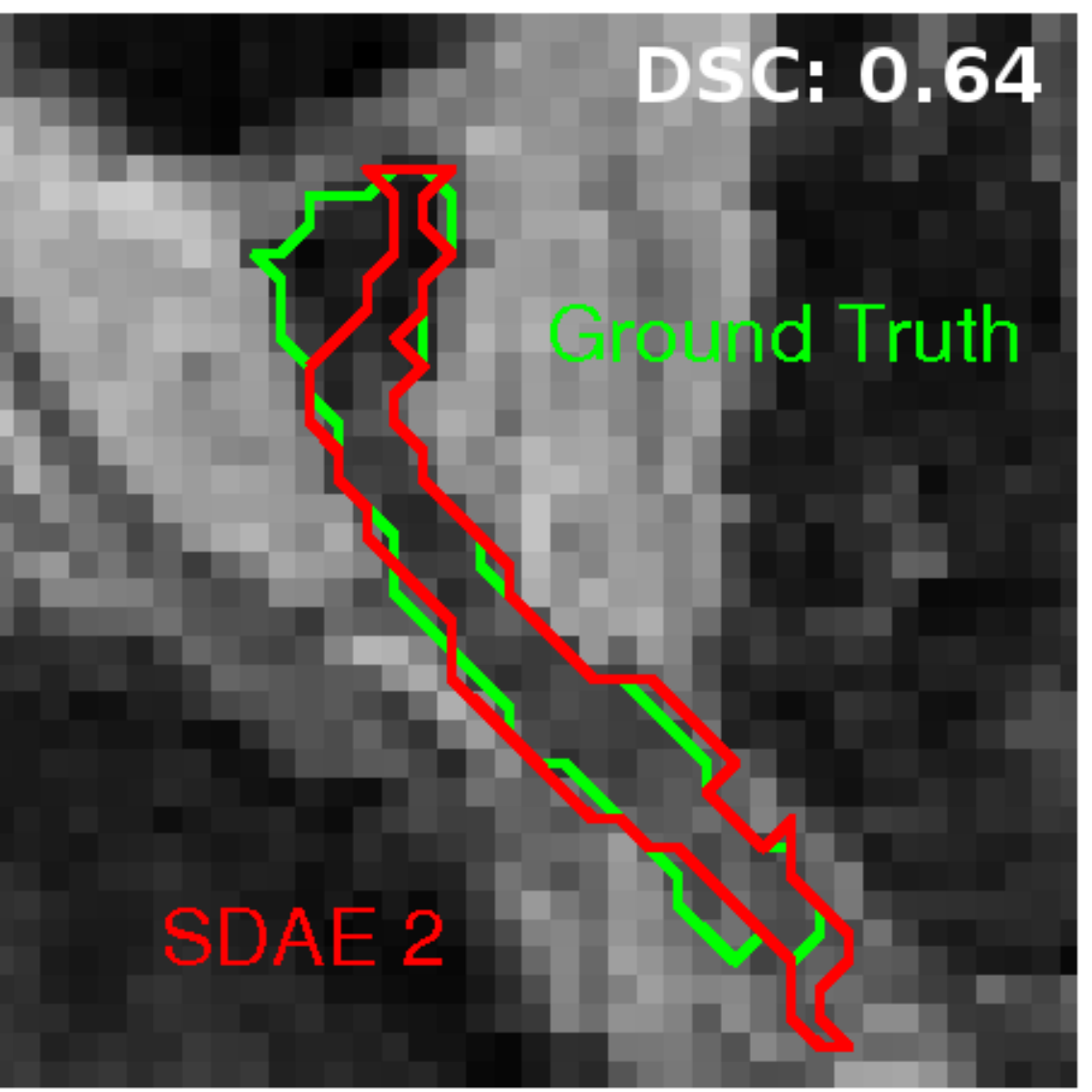}} \hspace{0.01em}
\subfloat{\includegraphics[width=0.21\linewidth]{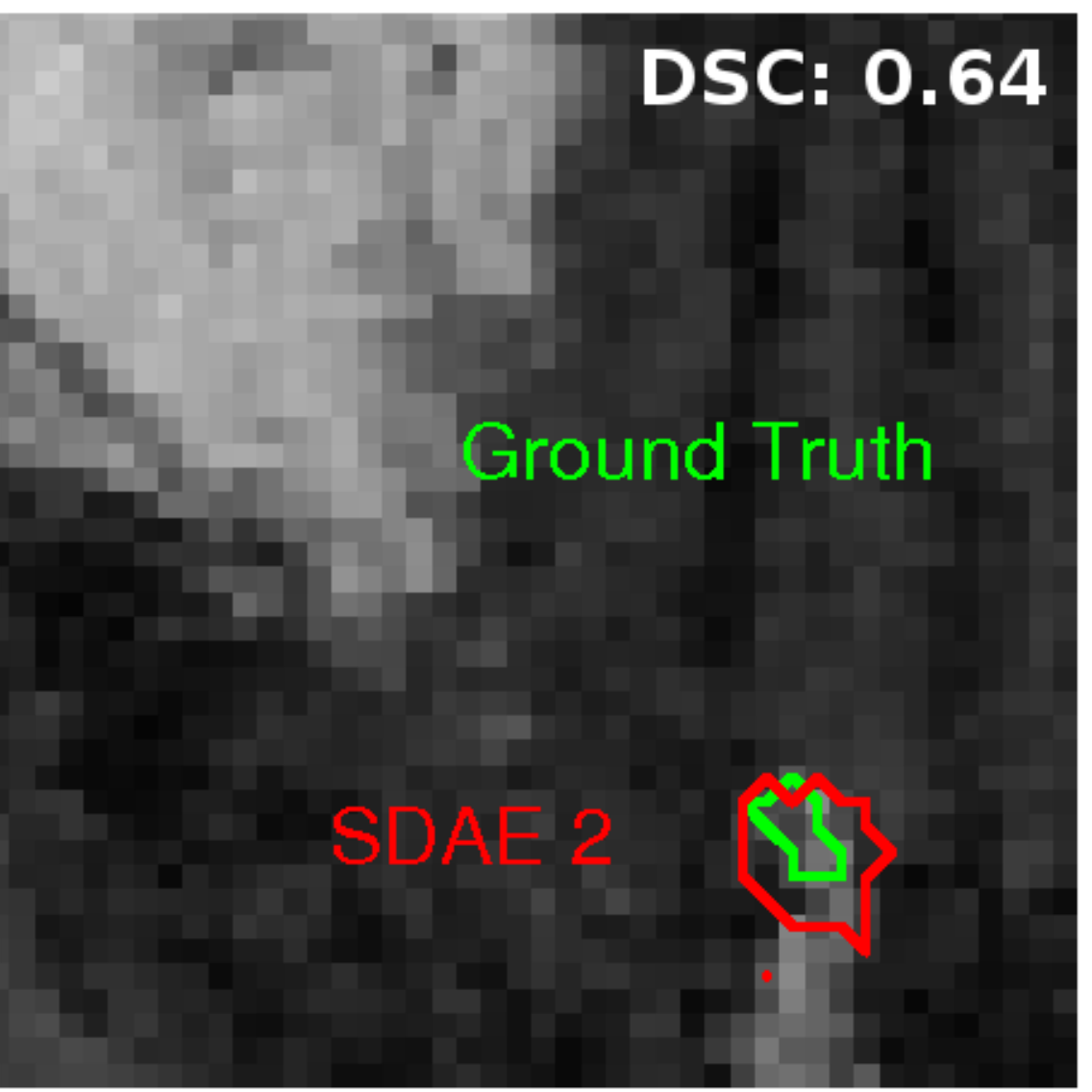}} \hspace{0.01em}
\subfloat{\includegraphics[width=0.21\linewidth]{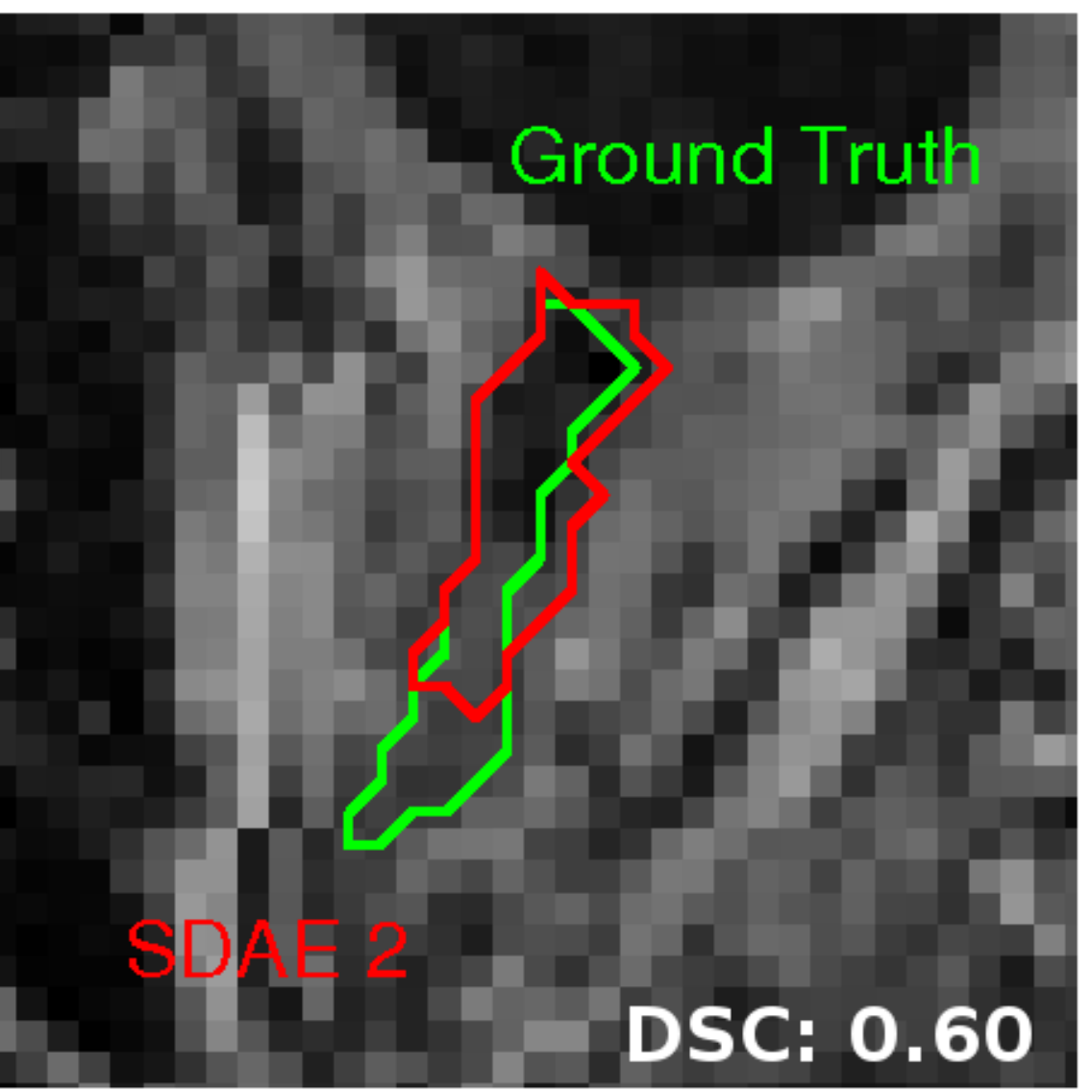}} \hspace{0.01em}
\subfloat{\includegraphics[width=0.21\linewidth]{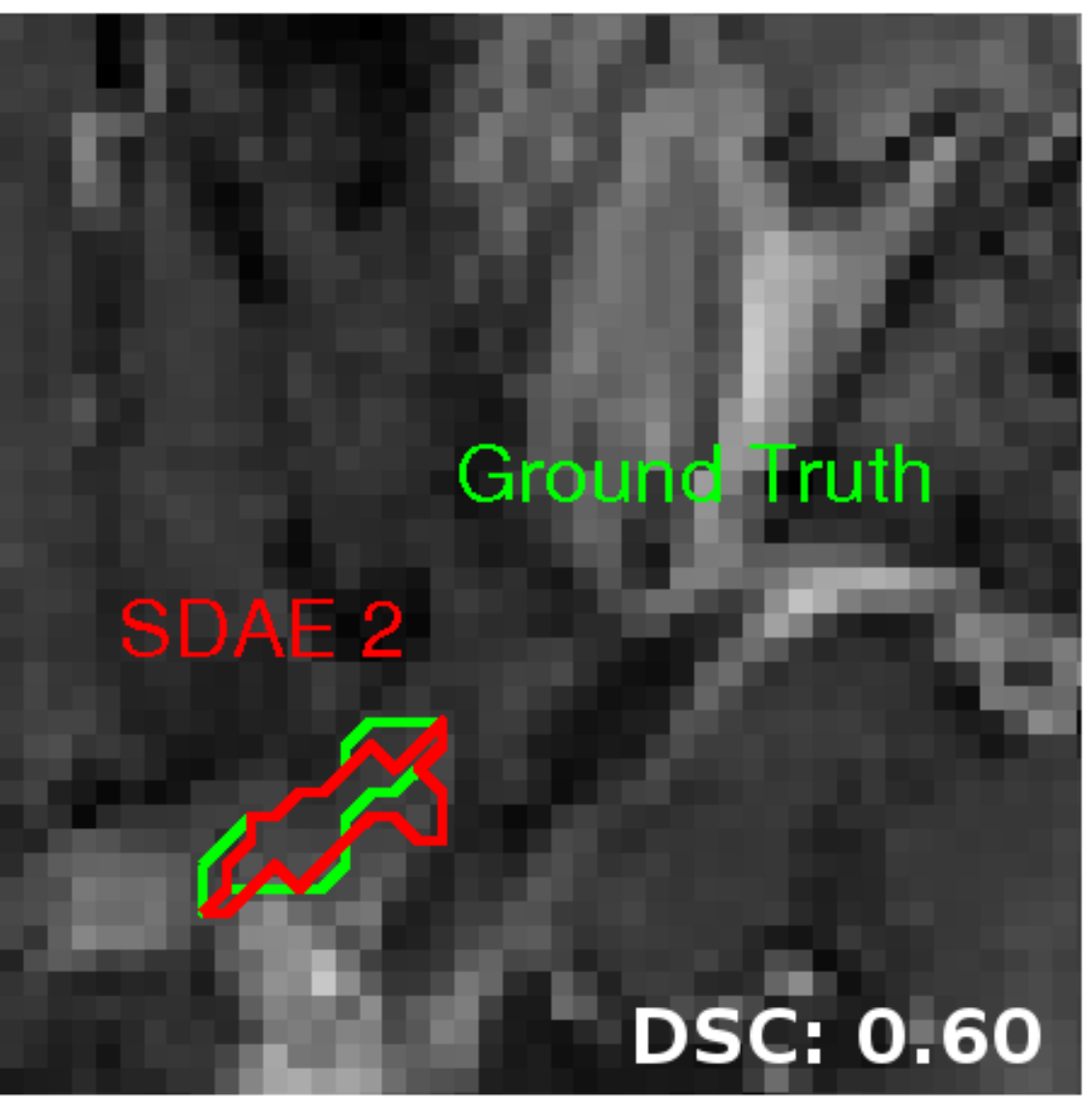}} 
\end{tabular}
\caption{Best and worst optic nerves segmentations generated by the proposed deep learning approach. While best segmentations are shown on the top, worst segmentations cases are shown on the bottom.}
\label{fig:BestWorstONSeg}
\end{figure}

It is important to note that similarity metrics are very sensitive in small organs. Differences in only few voxels can considerably increase or decrease comparison values. Therefore, we consider that having obtained DSC values higher than 0.7 in small OARs is very satisfactory, in addition with good values for the other metrics. Even in the worst cases, where DSC was above 0.55-0.60 for all the organs analyzed, the automatic contours can be considered as a good approximation of the reference. As example, Figure \ref{fig:BestWorstONSeg} shows the best and worst segmentation for both left and right optic nerves. 

In the context of structure delineation for radiation therapy, there exist a trade-off between preservation of structures of interest, need to sufficient treat tumor, and the ability to accurately deliver dose. Based on the results, we have demonstrated that the proposed approach can successfully address preservation of OARs, while allowing the PTV to be irradiated. Furthermore, segmentation is achieved in a fraction of time with respect to other presented approaches. We believe that its adoption in RTP might therefore facilitate the segmentation task.

Results obtained with the incorporation of the proposed features into the features vector to feed the deep network suggest that we are going in the good direction. Nevertheless, it is important to note that differences in data acquisition as well as differences in manual contours used as reference might compromise comparisons with other works. The lack of public datasets including the structures of interest makes also difficult comparison with other approaches.




\section{Conclusion}
\label{sec:conclusion}

We have proposed a deep learning based classification system to segment small organs at risk of the optic region in brain cancer. In addition to classical features widely employed in machine learning to segment brain structures, we have incorporated contextual features and textural features, leading to an augmented and enhanced features vector (AE-FV). Experimental results have shown that the proposed scheme achieve satisfactory results in terms of segmentation accuracy and processing time, with respect to the reference contours. Additionally, incorporation of proposed features yields to improvements on the segmentation with respect to classical features. This study has also shown how segmentation of some OARs in brain cancer can benefit from the synergy between hand-crafted features and deep learning representation.

\vspace{5mm} 
\textbf{Acknowledgments.} This project has received funding from the European Union’s Seventh Framework Programme for research, technological development and demonstration under grant agreement no PITN-GA-2011-290148.

\section*{References}

\bibliographystyle{unsrt} 

\end{document}